\documentclass[conference,compsoc]{IEEEtran}
\usepackage{amsfonts,amssymb}
\usepackage{graphicx}
\usepackage{grffile}
\usepackage{bbding}
\usepackage{booktabs}
\usepackage{multirow}
\usepackage{subfigure} 
\usepackage{wrapfig}
\usepackage{geometry}
\usepackage{xcolor}
\usepackage{hyperref}
\usepackage{amsmath}
\usepackage{amssymb}% http://ctan.org/pkg/amssymb
\usepackage{pifont}% http://ctan.org/pkg/pifont
\newcommand{\xmark}{\ding{55}}%
\ifCLASSOPTIONcompsoc
  \usepackage[nocompress]{cite}
\else
  \usepackage{cite}
\fi
\ifCLASSINFOpdf
\else
\fi
\begin{document}
% \title{Fast Online Object Tracking and Segmentation: A Unifying Approach}
\title{SiamMask: A Framework for Fast Online Object Tracking and Segmentation}
% \title{A Framework for Fast Online Object Tracking and Segmentation}

% \author{\IEEEauthorblockN{Weiming Hu and Qiang Wang}
% \IEEEauthorblockN{(National Laboratory of Pattern Recognition, Institute of Automation, Chinese Academy of Sciences, Beijing 100190) \\
% \{wmhu, qiang.wang\}@nlpr.ia.ac.cn} \\

% \IEEEauthorblockN{Li Zhang}
% \IEEEauthorblockN{(School of Data Science, Fudan University, Shanghai, 200433)\\
% lizhangfd@fudan.edu.cn} \\

% \IEEEauthorblockN{Luca Bertinetto}
% \IEEEauthorblockN{(Five AI Limited, Oxford OX1 3PJ, United Kingdom) \\
% luca.bertinetto@five.ai}\\

% \IEEEauthorblockN{Philip H.S. Torr}
% \IEEEauthorblockN{((Department of Engineering Science, University of Oxford, Oxford OX1 3PJ, United Kingdom) \\
% philip.torr@eng.ox.ac.uk
% }}

\author{Weiming Hu$^1$,Qiang Wang$^1$, Li Zhang$^2$, Luca Bertinetto$^3$, Philip H.S. Torr$^4$ \\
$^1$National Laboratory of Pattern Recognition, Institute of Automation, Chinese Academy of Sciences, Beijing 100190 \\
$^2$School of Data Science, Fudan University, Shanghai, 200433,
$^3$Five AI Limited, Oxford OX1 3PJ, United Kingdom \\
$^4$Department of Engineering Science, University of Oxford, Oxford OX1 3PJ, United Kingdom \\
\{wmhu, qiang.wang\}@nlpr.ia.ac.cn, lizhangfd@fudan.edu.cn, luca.bertinetto@five.ai, philip.torr@eng.ox.ac.uk

}


\newcommand{\drafty}[1]{\textcolor{orange}{#1}}
\newcommand*{\eg}{e.g.\@\xspace}
\newcommand*{\ie}{i.e.\@\xspace}
\maketitle
\begin{abstract}
In this paper we introduce SiamMask, a framework to perform both visual object tracking and video object segmentation, in real-time, with the same simple method.
We improve the offline training procedure of popular fully-convolutional Siamese approaches by augmenting their losses with a binary segmentation task.
Once the offline training is completed, SiamMask only requires a single bounding box for initialization and can simultaneously carry out visual object tracking and segmentation at high frame-rates.
Moreover, we show that it is possible to extend the framework to handle multiple object tracking and segmentation by simply re-using the multi-task model in a cascaded fashion.
Experimental results show that our approach has high processing efficiency, at around 55 frames per second. It yields real-time state-of-the-art  results on visual-object tracking benchmarks, while at the same time demonstrating competitive performance at a high speed for video object segmentation benchmarks.
% The project website is \url{http://www.robots.ox.ac.uk/~qwang/SiamMask}.
\end{abstract}

\begin{IEEEkeywords}
Object tracking, Video object segmentation, multiple object tracking, Siamese network.
\end{IEEEkeywords}

% \fi
\IEEEpeerreviewmaketitle

\section{Introduction}
\label{sec:introduction}
Tracking is a fundamental task in any video application requiring some degree of reasoning about objects of interest, as it allows to establish object correspondences between frames. 
It finds use in a wide range of scenarios such as automatic surveillance, vehicle navigation, video labelling, human-computer interaction and activity recognition.
Given the location of an arbitrary target of interest in the first frame of a video, the aim of \emph{visual object tracking} is to estimate its position in all the subsequent frames with the best possible accuracy~\cite{smeulders2013visual}.
For many applications, it is important that tracking can be performed \emph{online}, \emph{i.e.} while the video is streaming, which implies that the tracker should not make use of future frames to reason about the current position of the object~\cite{kristan2018sixth}.
This is the scenario portrayed by visual-object tracking benchmarks, which represent the target object with a simple axis-aligned (\cite{liang2015encoding,mueller2016benchmark,muller2018trackingnet,valmadre2018long}) or rotated~\cite{kristan2017visual,kristan2018sixth} bounding box.
Such a simple annotation helps to keep the cost of data labelling low; what is more, it allows a user to perform a quick and simple initialization of the target.
However, in the presence of complex movements and non-rigid deformations, bounding boxes are a very poor approximation of an object's contour, which can cause the erroneous inclusion of pixels belonging to the background in the representation.

Similar to object tracking, the task of \emph{video object segmentation} (VOS) requires estimating the position of an arbitrary target specified in the first frame of a video.
However, in this case the object representation consists of a binary segmentation mask expressing whether or not a pixel belongs to the target~\cite{perazzi2016benchmark}.
Such a detailed representation is more desirable for applications that require detailed pixel-level information, like video editing~\cite{perazzi2017video} and rotoscoping~\cite{miksik2017roam}.
Understandably, producing pixel-wise masks requires more computational resources than a simple bounding box.
As a consequence, VOS methods have been traditionally slow, often requiring several seconds per frame (\emph{e.g.}~\cite{wen2015jots,tsai2016video,perazzi2017learning,bao2018cnn}).
Recently, there has been a surge of interest in faster approaches~\cite{yang2018efficient,marki2016bilateral,oh2018fast,cheng2018fast,chen2018blazingly,jampani2017video,hu2018videomatch}.
However, even the fastest have not been able to operate in real-time.

\begin{figure}[t]
	\centering
	\includegraphics[width=\columnwidth]{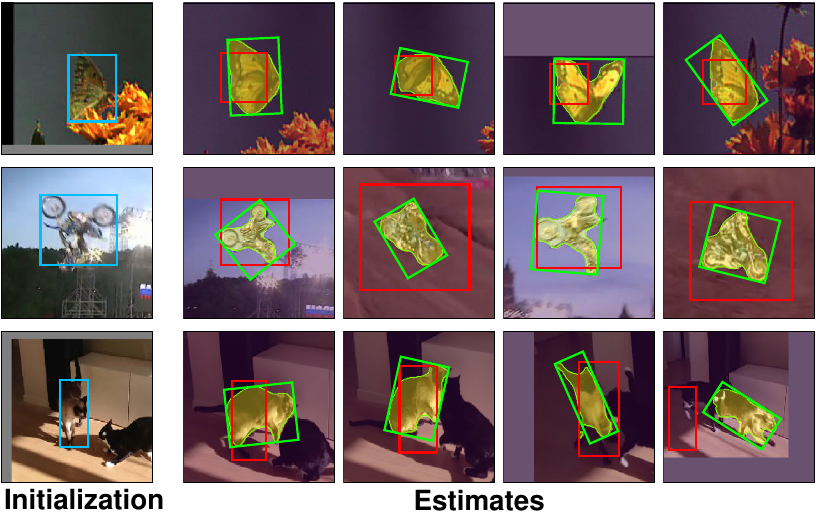}
	\caption{
    Our method addresses both tasks of visual tracking and video object segmentation to achieve high practical convenience.
    Like conventional object trackers such as~\cite{danelljan2017eco} (red), it relies on a simple bounding box initialization (blue) and operates online.
    However, SiamMask (green) is able to produce binary segmentation masks (out of which we can infer rotated bounding-boxes) that much more accurately describe the target object.}
    \label{fig:teaser}
\end{figure}

We aim at narrowing the gap between arbitrary object tracking and VOS by proposing \textit{SiamMask}, a simple multi-task learning approach that can be used to address \emph{both} problems.
Our method is motivated by the success of fast tracking approaches based on fully-convolutional Siamese networks~\cite{bertinetto2016fully} trained offline on millions of pairs of video frames (\emph{e.g.}~\cite{li2018high,zhu2018distractor,he2018towards,yang2018learning}) and by the recent availability of YouTube-VOS~\cite{xu2018youtube}, a large video dataset with pixel-wise annotations. 
We aim at retaining both the offline trainability and online speed of Siamese approches, while at the same time significantly refining their representation of the target object, which is limited to a simple axis-aligned bounding box.
To achieve this goal, we simultaneously train a fully-convolutional Siamese network on three tasks, each corresponding to a different strategy to establish correspondances between the target object and candidate regions in the new frames.
As in the work of Bertinetto \emph{et al.
}~\cite{bertinetto2016fully}, one task is to learn a measure of similarity between the target object and multiple candidates in a sliding window fashion.
The output is a dense response map which only indicates the location of the object, without providing any information about its spatial extent.
To refine this information, we simultaneously learn two further tasks: bounding box regression using a Region Proposal Network~\cite{ren2015faster,li2018high} and class-agnostic binary segmentation~\cite{o2015learning}.
Notably, the segmentation binary labels are only required during offline training to compute the segmentation loss and \emph{not} online during segmentation/tracking. 
In our proposed architecture, each task is represented by a different branch departing from a shared CNN and contributes towards a final loss, which sums the three outputs together.

Once trained, SiamMask solely relies on a single bounding box initialisation, operates online without updates and produces object segmentation masks and rotated bounding boxes at 55 frames per second (on a single consumer-grade GPU).
Despite its simplicity and fast speed, SiamMask establishes a new state-of-the-art for the problem of real-time object tracking.
Moreover, the \emph{same method} is also very competitive against VOS approaches (on multiple benchmarks), while being the fastest by a large margin.
This result is achieved with a simple bounding box initialisation (as opposed to a mask) and without adopting costly techniques often used by VOS approaches such as fine-tuning~\cite{maninis2018video,perazzi2017learning,bao2018cnn,voigtlaender2017online}, data augmentation~\cite{khoreva2017lucid,li2018video} and optical flow~\cite{tsai2016video,bao2018cnn,perazzi2017learning,li2018video,cheng2018fast}.

We further extend our multi-task framework to the problem of multiple object tracking and segmentation by adopting a trained SiamMask model within a two-stage cascade strategy.
Using multiple instances of SiamMask (one per target object), the first stage identifies a crop where the target object is likely to be located, and the second extracts an accurate pixel-wise mask.
As common in multiple-object tracking, the data association problem (where new target objects are mapped to existing tracks) is addressed with the Hungarian algorithm.
This overall strategy is fairly effective, and despite its simplicity it achieved the second place at the YouTube-VIS challenge~\cite{yang2019video}.

The rest of this paper is organized as follows.
Section~\ref{sec:relwork} briefly reviews related work on visual-object tracking and video-object segmentation.
Section~\ref{sec:background} presents an overview on fully-convolutional Siamese networks, which are at the basis of our work.
Section~\ref{sec:method} describes our proposed approach for tracking and segmentation, while Section~\ref{sec:MOT} explains how it can be extended to address the problem of multiple object tracking and segmentation. Section~\ref{sec:experiments} reports quantitative and qualitative experimental results for all the tasks considered on several popular benchmarks.
Finally, Section~\ref{sec:conclusion} concludes the paper.

\section{Related Work}
\label{sec:relwork}
To provide context for our work, we briefly discuss some of the most representative developments in visual-object tracking and video-object segmentation of the last few years.

\subsection{Visual object tracking}
Until recently, the most popular paradigm for tracking arbitrary objects was to train online a discriminative classifier from the ground-truth information provided in the first frame of a video, and then update the classifier online \cite{smeulders2013visual}.
One particularly popular and effective strategy was the use of the Correlation Filter~\cite{bolme2010visual}, a simple algorithm that allows to discriminate between the template of an arbitrary target and its 2D translations at high speeds thanks to its formulation in the Fourier domain.
Since the pioneering work of Bolme~\emph{et al.}, performance of Correlation Filter-based trackers has been notably improved with the adoption of multi-channel formulations~\cite{kiani2013multi,henriques2014high}, spatial constraints~\cite{kiani2015correlation,danelljan2015learning,lukezic2017discriminative,li2018learning} and deep features (\emph{e.g.}~\cite{danelljan2017eco,valmadre2017end}). 

In 2016, a different paradigm started gaining popularity~\cite{bertinetto2016fully,held2016learning,tao2016siamese}. 
Instead of learning a discriminative classifier online, these methods train offline a similarity function on pairs of video frames.
At test time, this function can be simply evaluated on a new video, once per frame.
In particular, evolutions of the fully-convolutional Siamese approach~\cite{bertinetto2016fully} considerably improved tracking performance by making use of region proposals~\cite{li2018high}, hard negative mining~\cite{zhu2018distractor}, ensembling~\cite{he2018towards,he2018twofold}, memory networks~\cite{yang2018learning}, multiple stage cascaded regression \cite{wang2019spm,fan2019siamese}, and anchor-free mechanism \cite{xu2020siamfc++,guo2020siamcar,chen2020siamese}
More recent developments improved the fully-convolutional framework under several different aspects.
Yang~\emph{et al.}~\cite{yang2020roam} recur to meta-learning to iteratively adjust the network's parameters during tracking via the use of an offline-trained recurrent neural network.
Cheng~\emph{et al.}~\cite{cheng2021learning} focuses on the challenging issue brought by ``distractors'' during tracking by learning explicitly learning a ``relation detector'' to discriminate them from the background.
Guo~\emph{et al.}~\cite{guo2021graph} propose to enhance the Siamese network with a graph attention mechanism to establish correspondences between the features of target and search area.
Zhou~\emph{et al.}~\cite{zhou2021saliency} focus on mining the most salient regions of the tracked objects to increase the discriminative power of the trained model.
Yan~\emph{et al.}~\cite{yan2021lighttrack} utilize Neural Architecture Search to prune the large space of Siamese-based architecture and find the best performing or the most efficient in terms of FLOPs.
Furthermore, several works~\cite{jia2020robust,liang2020efficient,nakka2020temporally,yan2020cooling,guo2020spark} are concerned with the negative implications of adversarial attacks applied to tracking systems, and address them with several techniques from the robustness literature (\emph{e.g.} adversarial training).

A recently popular trend is the one of using self-supervised approaches for visual-object tracking.
Wang~\emph{et al.}~\cite{wang2021unsupervised} propose a Siamese Correlation Filter-based network trained using pseudo-labels obtained by running a tracker back and forth on a video to obtain stable trajectories.
Yuan~\emph{et al.}~\cite{yuan2020self} exploit cycle-consistency for representation learning.
Zheng~\emph{et al.}~\cite{zheng2021learning} learn a Siamese network in an unsupervised way by mining moving objects in video via optical flow. 
Similarly, Sio~\emph{et al.}~\cite{sio2020s2siamfc} manage to learn a Siamese network in an unsupervised way by extracting both exemplar and search image from the same frame, while employing data augmentation to not make the problem trivial. Finally, Wu~\emph{et al.}~\cite{wu2021progressive} learn a foreground/background discriminator in an unsupervised way using contrasting learning.
This is then used to discover corresponding patches throughout a video to learn the tracking model.

The above trackers use a rectangular bounding box both to initialise the target \emph{and} to estimate its position in the subsequent frames.
Despite its convenience, a simple rectangle often fails to properly represent an object, as it is evident in the examples of Figure~\ref{fig:teaser}.
This motivated us to propose a tracker able to produce binary segmentation masks while still only relying on a quick-to-draw bounding box initialization.

\subsection{Video object segmentation (VOS)}

Benchmarks for arbitrary object tracking (\emph{e.g.}~\cite{smeulders2013visual,kristan2016novel}) assume that trackers receive input frames in a sequential fashion.
This aspect is generally referred to with the attributes \emph{online} or \emph{causal}~\cite{kristan2016novel}.
Moreover, methods are often focused on achieving a speed that exceeds the one of typical video frame rates (around 25 to 30 frames per second).
Conversely, VOS algorithms have been traditionally more concerned with an accurate representation of the object of interest~\cite{perazzi2017video,perazzi2016benchmark}.

In order to exploit consistency between video frames, several methods propagate the supervisory segmentation mask of the first frame to the temporally adjacent ones via graph labeling approaches (\emph{e.g.}~\cite{wen2015jots,perazzi2015fully,tsai2016video,marki2016bilateral,bao2018cnn,zhang2020transductive}). 
In particular, Bao \emph{et al.}~\cite{bao2018cnn} recently proposed a very accurate method that makes use of a spatio-temporal MRF in which temporal dependencies are modelled by optical flow, while spatial dependencies are expressed by a CNN.
Another popular strategy is to process video frames independently (\emph{e.g.}~\cite{maninis2018video,perazzi2017learning,voigtlaender2017online}), similarly to what happens in most tracking approaches.
For example, in OSVOS-S Maninis \emph{et al.}~\cite{maninis2018video} do not make use of any temporal information.
They rely on a fully-convolutional network pre-trained for classification and then, at test time, they fine-tune it using the ground-truth mask provided in the first frame.
MaskTrack~\cite{perazzi2017learning} instead is trained from scratch on individual images, but it does exploit some form of temporality at test time by using the latest mask prediction and optical flow as additional input to the network.

Aiming towards the highest possible accuracy, at test time VOS methods often feature computationally intensive techniques such as fine-tuning~\cite{maninis2018video,perazzi2017learning,bao2018cnn,voigtlaender2017online}, data augmentation~\cite{khoreva2017lucid,li2018video} and optical flow~\cite{tsai2016video,bao2018cnn,perazzi2017learning,li2018video,cheng2018fast}.
Therefore, these approaches are generally characterized by low frame rates and the inability to operate online.
For example, it is not uncommon for methods to require minutes~\cite{perazzi2017learning,cheng2017segflow} or even hours~\cite{tsai2016video,bao2018cnn} for videos that are just a few seconds long, like the ones of the DAVIS benchmark~\cite{perazzi2016benchmark}.
Recently, there has been an increasing interest in the VOS community towards \emph{faster} methods~\cite{marki2016bilateral,oh2018fast,cheng2018fast,chen2018blazingly,jampani2017video,hu2018videomatch}.
Two notable fast approaches with a performance competitive with the state of the art are OSMN~\cite{yang2018efficient} and RGMP~\cite{oh2018fast}.
The former uses a meta-network ``modulator'' to quickly adapt the parameters of a segmentation network during test time, while the latter does not use any fine-tuning and adopts an encoder-decoder Siamese architecture trained in multiple stages.
Both these methods run at less than 10 frames per second, which does not make them suitable for real-time applications.
 
\subsection{Tracking and segmentation}
Interestingly, in the past it was not uncommon for online trackers to produce a very coarse binary mask of the target object (\emph{e.g.}~\cite{comaniciu2000real,perez2002color,bibby2008robust,possegger2015defense}).
In modern times, faster and online-operating trackers have typically used rectangular bounding boxes to represent the target object, while to be able to produce accurate masks researchers have often forego speed and online operability, as we saw in the previous section.

Few notable exceptions exists, some of which are very recent and have been published after the conference version of this paper.
Yeo~\emph{et al.}~\cite{yeo2017superpixel} proposed a super pixel-based tracker that is able to operate online and produce binary masks for the object starting from a bounding box initialization.
However, the fastest variant of this tracker runs at 4 frames per second. When the CNN features were used, its speed is decreased by 40 times
Perazzi~\emph{et al.}~\cite{perazzi2017learning} and Ci~\emph{et al.}~
\cite{ci2018video} propose video object segmentation methods which, like us, can be initialized using a simple axis-aligned rectangle in the first frame while also outputting a mask at each frame.
Yan~\emph{et al.}~\cite{yan2021alpha} adopts a pixel-wise correlation layer and an auxiliary mask head to improve tracking performance.
However, their method requires online learning of the network parameters, which restricts its practical applications.
Lukezic~\emph{et al.}~\cite{lukezic2020d3s} propose an approach that handles both tracking and segmentation by encoding the target object with two discriminative models capturing complementary properties: one is adaptive but only considers Euclidean motions, the other instead accounts for non-rigid transformations.
These methods require online learning during tracking, which can impact their speed.

\section{Fully-convolutional Siamese Networks}
\label{sec:background}
To allow online operability and fast speed, we adopt the fully-convolutional Siamese framework~\cite{bertinetto2016fully}, considering both SiamFC~\cite{bertinetto2016fully,valmadre2017end} and SiamRPN~\cite{li2018high} as the method to start from.
We first introduce them in Section~\ref{sec:siamfc} and~\ref{sec:siamrpn} and then describe our approach in Section~\ref{sec:method}.

\subsection{SiamFC}
\label{sec:siamfc}

Bertinetto \emph{et al.}~\cite{bertinetto2016fully} proposed to use, as a fundamental building block of a tracking system, an offline-trained fully-convolutional Siamese network that compares an exemplar image \textbf{z} against a larger search image \textbf{x} to obtain a dense response map (where the location with the highest response can be used to infer the location of the exemplar in the search image).
\textbf{z} and \textbf{x} are, respectively, a crop of size $w{\times}h$ centered on the target object  and a larger crop centered on the last estimated position of the target.
The two inputs are processed by the same CNN $f_{\theta}$, yielding two feature maps that are cross-correlated:
\begin{equation}\label{eq:cross}
    \textbf{g}_{\theta}(\textbf{z},\textbf{x}) = f_{\theta}(\textbf{z}) \star f_{\theta}(\textbf{x}).
\end{equation}
In this paper, we refer to each spatial element of the response map as \emph{response of a candidate window} (\textbf{RoW}), where $g_{\theta}^{n}(\textbf{z},\textbf{x})$ encodes the \textit{similarity} between the examplar \textbf{z} and the $n$-th candidate window in \textbf{x}.
For SiamFC, the goal is for the maximum value of the response map to correspond to the target location in the search area \textbf{x}.
In order to allow each RoW to encode richer information about the target object, as we will see later, we replace the simple cross-correlation of Eq.~\ref{eq:cross} with depth-wise cross-correlation~\cite{bertinetto2016learning,Kaiser2018} and produce a multi-channel response map.

SiamFC is trained offline on millions of video frames with the logistic loss~\cite[Section 2.2]{bertinetto2016fully}.
% , which we refer to as $\mathcal{L}_{sim}$.
Let $y^n \in \{+1, -1\}$ be the ground-truth label for the RoW of position $n$ in the grid $\mathcal{D}$ of the response map.
The logistic loss is defined as:
\begin{equation}
    \mathcal{L}_{sim}=\frac{1}{|\mathcal{D}|}\sum_{n \in \mathcal{D}}log(1+exp(-y^ng^n_{\theta}(\textbf{z},\textbf{x})).
\end{equation}

\subsection{SiamRPN}
\label{sec:siamrpn}

Li \emph{et al.}~\cite{li2018high} considerably improved the tracking accuracy of SiamFC by relying on a region proposal network (RPN)~\cite{ren2015faster,feichtenhofer2017detect}, which allows to estimate the target location with a bounding box of variable aspect ratio.
In particular, in SiamRPN each RoW encodes a set of $k$ anchor box proposals and corresponding object/background scores.
Therefore, SiamRPN is able to output box predictions (with a regression branch) in parallel with object/background classification scores.

Using the nomenclature and formulation from~\cite{li2018high}, let us assume a total of $k$ anchor boxes.
Convolutional layers are used to obtain the two features $[f(\textbf{z})]_{cls}$ (for the classification branch) and $[f(\textbf{z})]_{reg}$ (for the regression branch) from the feature map $f_{\theta}(\textbf{z})$.
Their number of channels depends on the number of anchors and increases, respectively, of $2k$-times and $4k$-times w.r.t. $f_{\theta}(\textbf{z})$.
The correlation between \textbf{z} and \textbf{x} on the classification branch is obtained by
\begin{equation}
    \mathbb{G}^{cls}_{|\mathcal{D}|\times 2k}=[f(\textbf{x})]_{cls} \star [f(\textbf{z})]_{cls},
\end{equation}
while the correlation on the regression branch by
\begin{equation}
    \mathbb{H}^{reg}_{w\times h \times 4k}=[f(\textbf{x})]_{reg} \star [f(\textbf{z})]_{reg}.
\end{equation}
This way, each spatial location in $\mathbb{G}$ and $\mathbb{H}$ has a ``depth'' of $2k$ and $4k$ channels respectively.
In other words, for each anchor from the RPN module, the network produces two multi-channel response maps:
\begin{itemize}
    \item A two-channel output for object/background classification, with ``scores'' representing corresponding location in the original map.
    \item A four-channels output for bounding-box regression, representing the center $(x, y)$ distance and the width and height $(w, h)$ difference between the anchor and corresponding ground truth.
\end{itemize}

Offline, the classification branch is trained using the cross-entropy loss $\mathcal{L}_{score}$ \cite{li2018high}:
\begin{equation}\label{eq:siamfc_loss}
    \mathcal{L}_{score}=\frac{1}{k|\mathcal{D}|}\sum_{n=1}^{|\mathcal{D}|}\sum_{i=1}^{k}-[y_{n,i}log(p_{n,i})+(1-y_{n,i})log(1-p_{n,i})],
\end{equation}
where $p_{n,i}$ is the output of the classification branch for the $i$-th anchor of the $n$-th RoW, while the regression branch is trained using the smooth $L_1$ loss $\mathcal{L}_{reg}$ with normalized coordinates.
Let $A_x$, $A_y$, $A_w$, and $A_h$ denote the coordinates of the center point and the width and height of an anchor box.
Let $T_x$, $T_y$, $T_w$, and $T_h$ denote the central coordinates, width, and height of the ground-truth box.
The normalized distance between an anchor and the ground-truth box is defined as
\begin{equation}
    \delta[0]=\frac{T_x-A_x}{A_w},\delta[1]=\frac{T_y-A_y}{A_h},\delta[2]=ln\frac{T_w}{A_w},\delta[3]=\frac{T_h}{A_h}.
\end{equation}
The smooth $L_1$ loss for $x$ is
\begin{equation}
    smooth_{L_1}(x,\beta)=\left\{
    \begin{array}{rcl}
         0.5\beta ^2 x^2 & & |x| < \frac{1}{\beta ^2}  \\
         |x|-\frac{1}{2\beta ^2} & & |x| \geq \frac{1}{\beta^2},
    \end{array}
    \right.
\end{equation}
where $\beta$ is a hyperparameter that needs tuning.
Let $\{q[j]\}^3_{i=0}$ be the outputs of the four channels of the regression branch for an anchor.
The loss $\mathcal{L}_{reg}$ for the anchor regression is defined as:
\begin{equation}
    \mathcal{L}_{reg}=\frac{1}{2k|\mathcal{D}|}\sum_{n=1}^{|\mathcal{D}|}\sum_{i=1}^k \sum_{j=0}^3 (y_i^n+1)smooth_{L_1}(\delta[j]-q[j], \beta),
\end{equation}
where $y_i^n$ is the ground-truth label for the $i$-th anchor of the $n$-th RoW. 

\section{SiamMask}
\label{sec:method}
Unlike existing tracking methods that rely on low-fidelity object representations, we argue the importance of producing per-frame binary segmentation masks.
To this aim we show that, besides similarity scores and bounding box coordinates, it is possible for the RoW of a fully-convolutional Siamese network to also encode the information necessary to produce a pixel-wise binary mask.
This can be achieved by extending existing Siamese trackers with an extra branch and loss.
In the following subsections, we describe the multi-branch network architecture (\ref{sec:multibranch}), the strategy to obtain and refine a mask representation (\ref{sec:mask_representation}), the loss function (\ref{sec:siammask_loss}), and how we obtain a bounding-box from a mask (\ref{sec:box_generation}).

\subsection{Multi-branch network architecture}
\label{sec:multibranch}
\begin{figure}[t]
	\centering
	\includegraphics[width=\columnwidth]{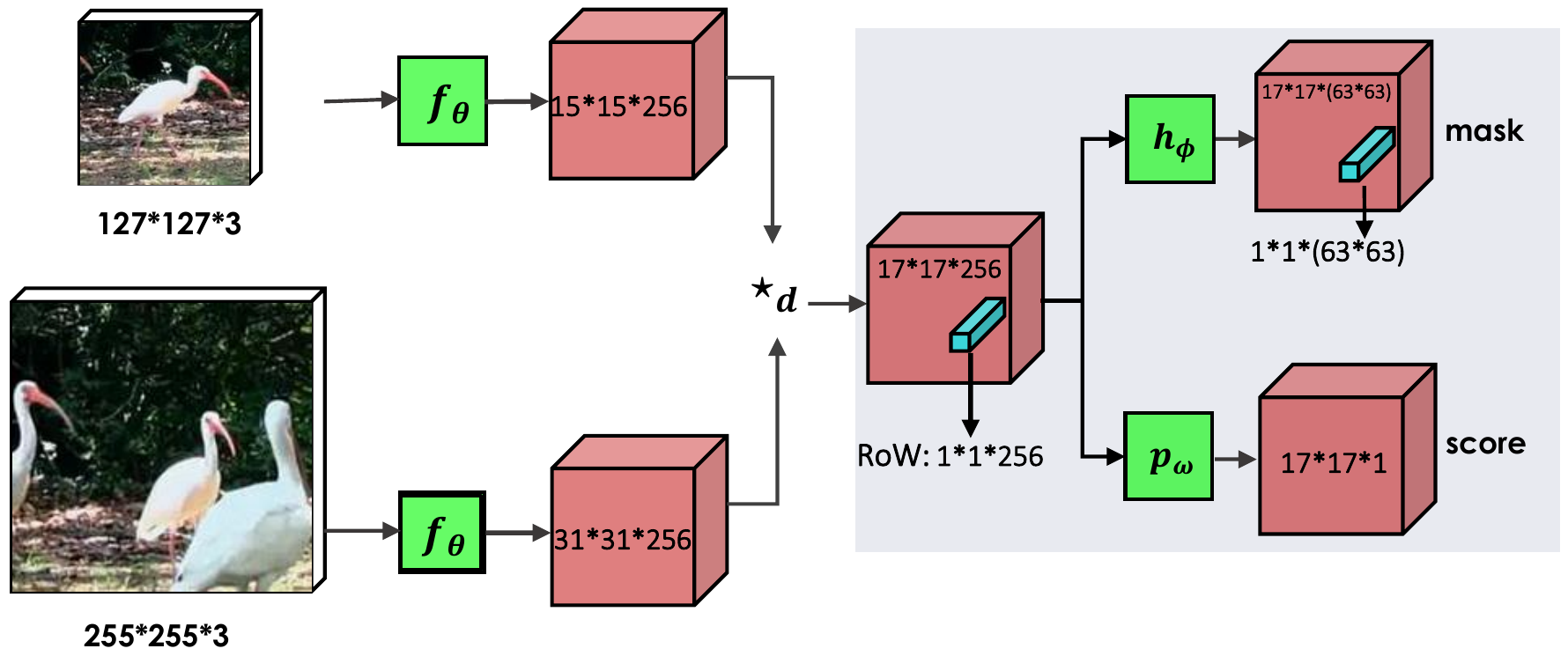}
	\caption{Schematic illustration of the two-branch variant of SiamMask, based on~\cite{bertinetto2016fully}. $\star_d$ denotes depth-wise cross correlation.}
	\label{fig:siammask_2b}
\end{figure}

\begin{figure}[t]
	\centering
	\includegraphics[width=\columnwidth]{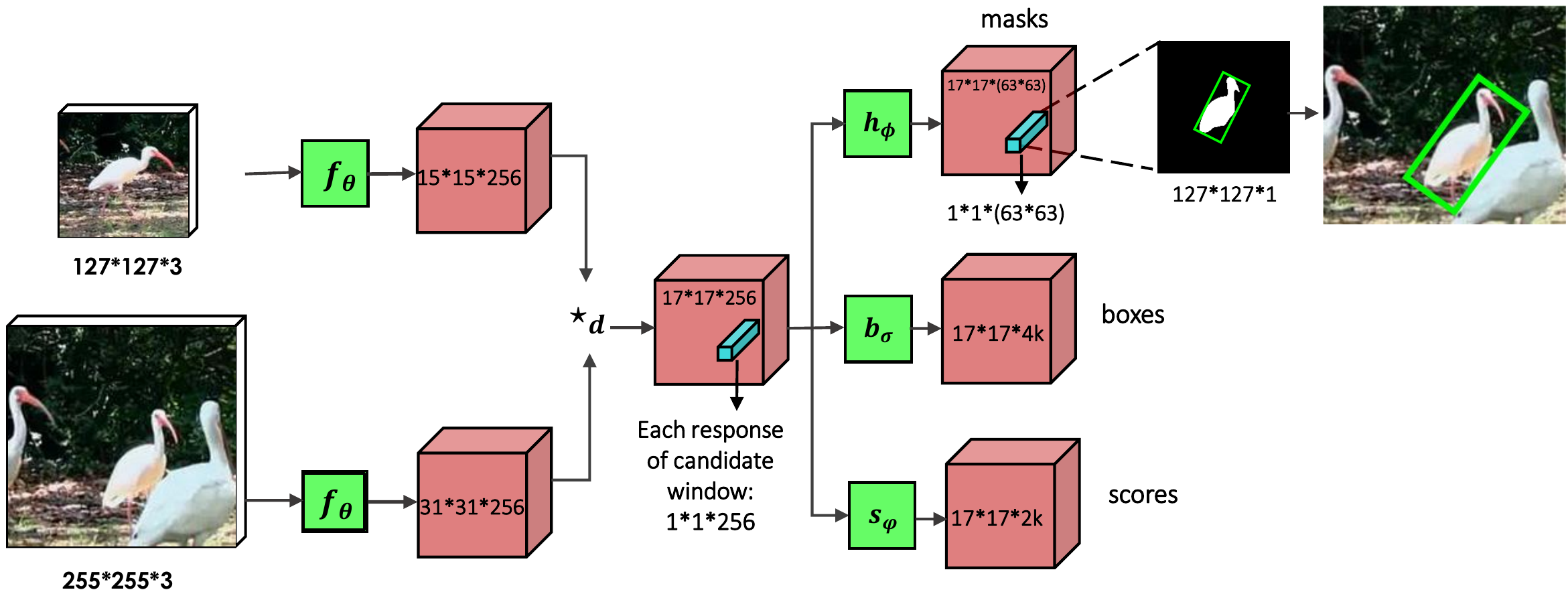}
	\caption{Schematic illustration of the three-branch variant of SiamMask, based on~\cite{li2018high}.}
	\label{fig:siammask_3b}
\end{figure}

We augment the architectures of SiamFC~\cite{bertinetto2016fully} and SiamRPN~\cite{li2018high} by adding our segmentation branch to obtain the two-branch and three-branch variants of the proposed Siamese mask network (SiamMask), illustrated in Fig.~\ref{fig:siammask_2b} and Fig.~\ref{fig:siammask_3b}.
In the two-branch variant, the branch $p_{\omega}$ is tasked with discriminating each RoW between target object or background, while the branch $h_{\phi}$ outputs one segmentation mask per RoW.
In addition to these, the three-branch variant also employs a box-regression branch $b_{\sigma}$ like in SiamRPN.

Importantly, in order to allow each RoW to encode richer information about the target object, we replace the simple cross-correlation ``$\star$'' of Eq.~\ref{eq:cross} with depth-wise cross-correlation (see \emph{e.g.} ~\cite{bertinetto2016learning}) ``$\star_d$'' and produce multi-channel response maps.
Then, the output of the depth-wise cross-correlation $g_{\theta}^n$ is a vector with $d$ channels (in the illustrative example of Fig.~\ref{fig:siammask_2b} it has size $1{\times}1{\times}256$).
Note that, for the classification branch, the multi-channel output is mapped to a single channel response map by a convolutional layer.
The logistic loss is then defined by replacing $g_{\theta}^n(\textbf{z},\textbf{x})$ from~\ref{eq:siamfc_loss} with $p_{\omega}(g_{\theta}^n(\textbf{z},\textbf{x}))$.

In the segmentation branch, we predict $w{\times}h$ binary masks $\textbf{m}^n$ (one for each RoW $n$) using a simple two-layers neural network $h_{\phi}$ with learnable parameters $\phi$:
\begin{equation}\label{eq:mask}
    \textbf{m}^n=h_{\phi}(g_{\theta}^n(\textbf{x}, \textbf{z})).
\end{equation}
From Eq.~\ref{eq:mask} we can see that the mask prediction is a function of \emph{both} the search are \textbf{x} and the exemplar image \textbf{z}.
This way, \textbf{z} can be used as a guide to ``prime'' the segmentation:
given a different reference image, the network will produce a different segmentation mask, as we will see in the experimental section.

\subsection{Mask representation and refinement}
\label{sec:mask_representation}

\begin{figure*}[t]
	\centering
	\includegraphics[width=\textwidth]{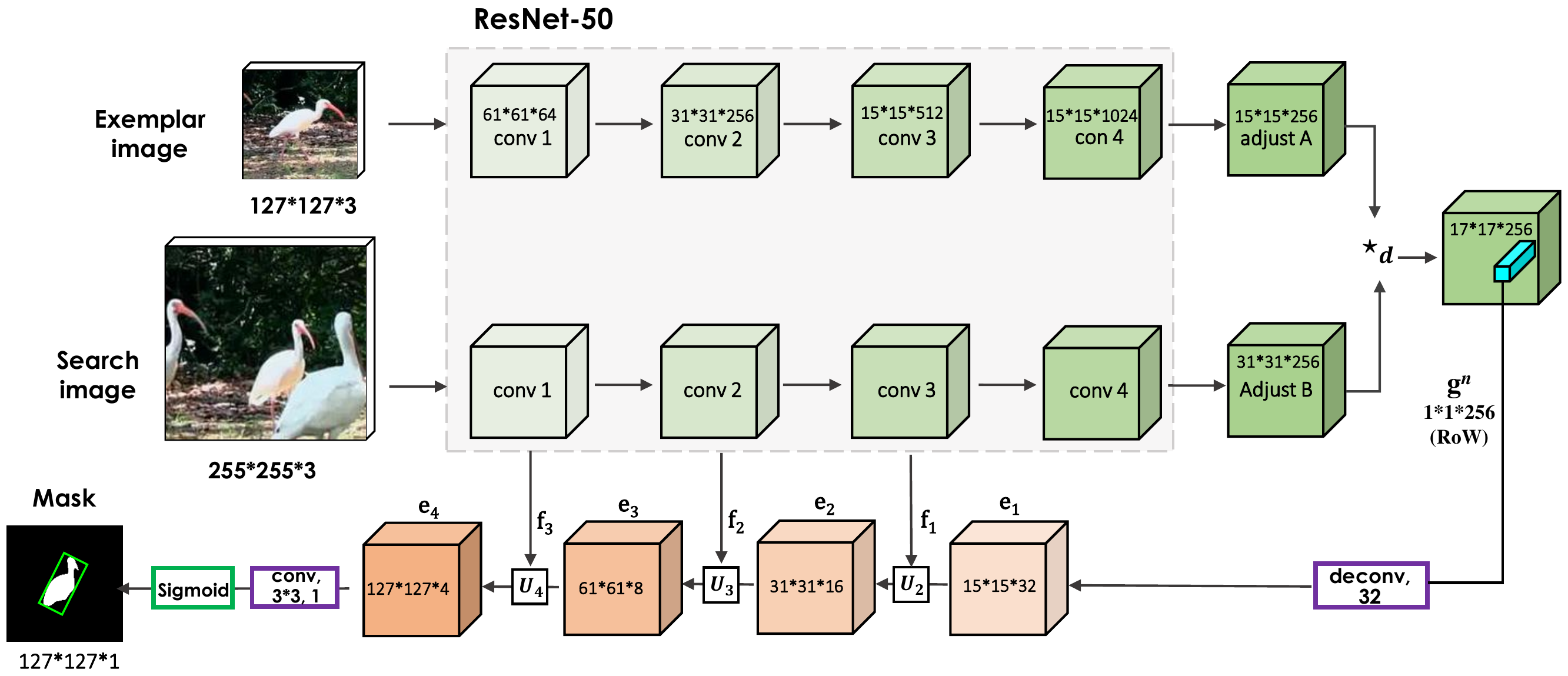}
	\caption{Schematic illustration of the stacked refinement modules. For a more detailed version of refinement modules, see Fig.~\ref{fig:refinement_module}.}
	\label{fig:arch_refinement_modules}
\end{figure*}

In contrast to semantic segmentation methods in the style of FCN~\cite{long2015fully} and Mask R-CNN~\cite{he2017mask}, which maintain explicit spatial information throughout the network, our approach follows the spirit of DeepMask~\cite{o2015learning} and SharpMask~\cite{pinheiro2016learning} and generates masks starting from a flattened representation of the object.
In particular, in our case this representation corresponds to one of the $17{\times}17$ RoWs produced by the depth-wise cross-correlation between $f_\theta(\textbf{z})$ and $f_\theta(\textbf{x})$.
Importantly, the network $h_\phi$ of the segmentation task is composed of two $1{\times}1$ convolutional layers, one with 256 and the other with $63^2$ channels (Figure~\ref{fig:siammask_2b}).
This allows every pixel classifier to utilise information contained in the entire RoW and thus to have a complete view of its corresponding candidate window in $\textbf{x}$.
With the aim of producing a more accurate object mask, we follow the strategy of~\cite{pinheiro2016learning}, which combines low and high resolution features using multiple~\textit{refinement} modules made of upsampling layers and skip connections.

Fig.~\ref{fig:arch_refinement_modules} represents a more detailed illustration of our architecture, which explicitly shows the stack of refinement modules for generating the final mask.
Exemplar and search image are processed by the same network, and their features (depth-wise) cross-correlated to obtain the features $\textbf{g}_{\theta}(\textbf{x}, \textbf{z})$, where we refer to the $n$-th RoW with $g^n_{\theta}(\textbf{x}, \textbf{z}) \in \mathbb{R}^{1\times 1 \times d}$.
Let $\textbf{f}_1$, $\textbf{f}_2$, and $\textbf{f}_3$ be the feature maps extracted from the third, second, and first layers in the Siamese network for the $n$-th RoW in \textbf{x}.
Deconvolution is carried out on $g^n_{\theta}(\textbf{x}, \textbf{z})$ to obtain the $n$-th RoW’s segmentation representation $\textbf{e}_1\in \mathbb{R}^{m_1\times m_1 \times k_1}$ at a relatively low resolution ($1<k_1<d, m_1>1$).
In the refinement module $U_2$, the mask representation $\textbf{e}_1$ and the high layer feature map $\textbf{f}_1$ are combined to obtain, by upsampling, a mask representation $\textbf{e}_2$ with a higher resolution than $\textbf{e}_1$: $\textbf{e}_2=U_2(\textbf{e}_1,\textbf{f}_1) \in \mathbb{R}^{m_2\times m_2 \times k_2}$ ($m_2>m_1, k_2<k_1$).
Analogously, modules $U_3$ and $U_4$ produce representations with increasingly higher resolutions.
Beside obtaining higher resolution feature maps, this procedure allows to use complementary information from layers at different ``depths''.

\begin{figure}[t]
	\centering
	\includegraphics[width=\columnwidth]{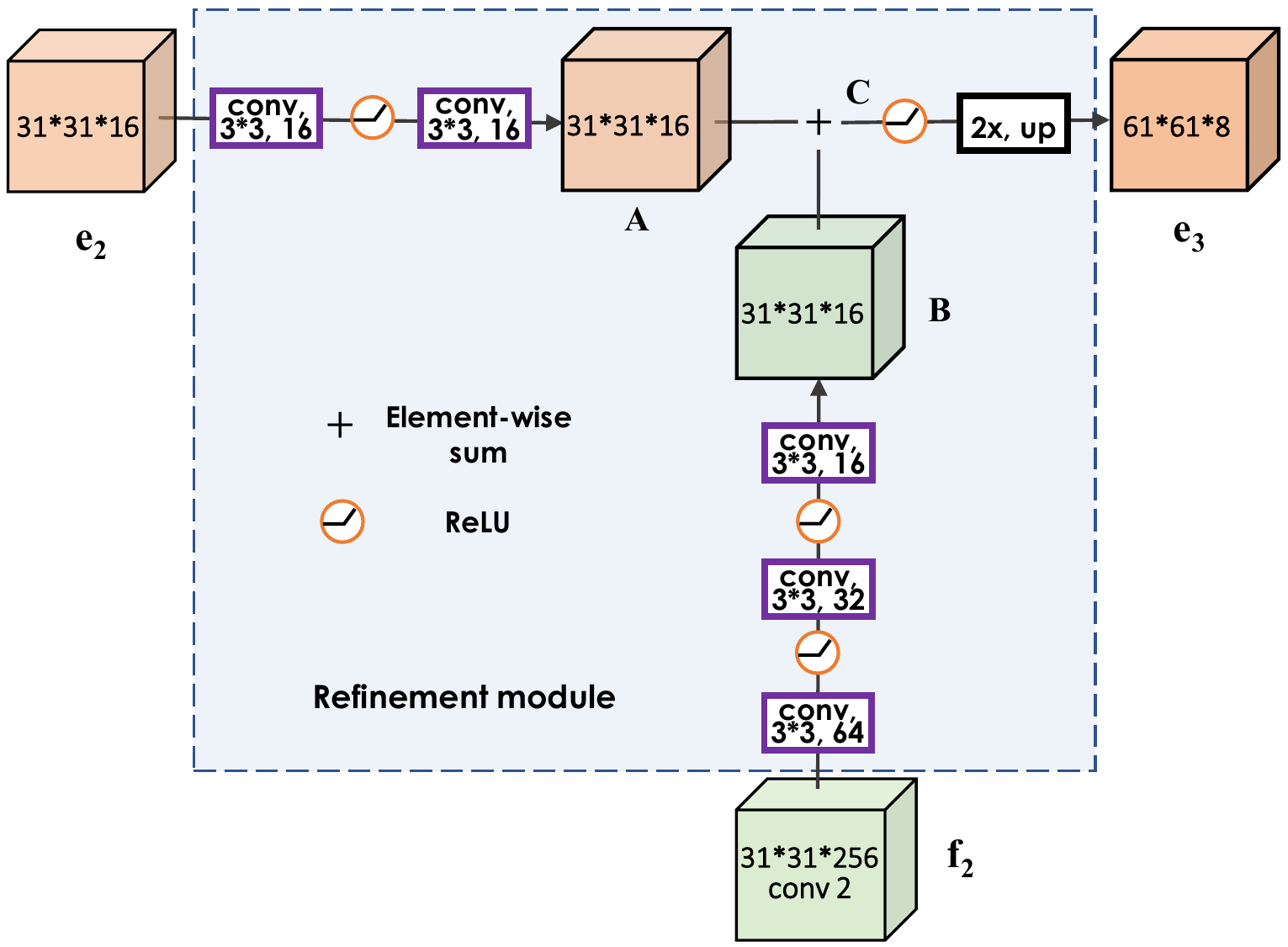}
	\caption{Schematic illustration of one of the three refinement modules ($U_3$).}
	\label{fig:refinement_module}
\end{figure}

Fig.~\ref{fig:refinement_module} shows the structure of $U_3$ as example of a refinement module.
The mask representation $\textbf{e}_2$ is used to obtain a new mask representation \textbf{A} via two convolutional layers and a non-linear layer.
The feature map $\textbf{f}_2$ is then used to output a new feature \textbf{B} with the same size as \textbf{A} via three convolutional layers and two non-linear layers.
The sum of \textbf{A} and \textbf{B} then produces a new mask \textbf{C}.
Finally, one last non-linear layer is used to produce a new, $2\times$-upscaled mask representation $\textbf{e}_3$.

\subsection{Loss function}
\label{sec:siammask_loss}
We define the loss function for the segmentation branch and combine it with the loss functions of the other branches.

\smallskip\noindent\textbf{Loss for the segmentation branch.}
During training, each RoW is labelled with a ground-truth binary label $y_{n} \in \{\pm 1\}$ and also associated with a pixel-wise ground-truth mask $c_n$ of size $w{\times}h$.
Let $c^{ij}_{n} \in \{\pm 1\}$ denote the label corresponding to pixel $(i,j)$ of the object mask in the $n$-th candidate RoW.
The loss function $\mathcal{L}_{mask}$ (Eq.~\ref{eq:loss}) for the mask prediction task is a binary logistic regression loss over all RoWs:

\begin{equation}\label{eq:loss}
		\mathcal{L}_{mask}(\theta,~\phi) =  \sum_{n} (\frac{1+y_{n}}{2wh}\sum_{ij} \log (1 + e^{-c^{ij}_{n}m_{n}^{ij}}
		)),
\end{equation}
where $m_{ij}^n$ is an element in $\textbf{m}^n$ defined in Eq.~\ref{eq:mask}.
Thus, the classification layer of $h_{\phi}$ consists of $w{\times}h$ classifiers, each indicating whether a given pixel belongs to the object in the candidate window or not.
Note that $\mathcal{L}_{mask}$ is considered only for positive RoWs (\emph{i.e.} with $y_{n}=1$).
Given the high number of negative samples, considering them would unbalance the loss.
We experimented with a weighted loss, so that negative and positive samples would hold the same importance, but it showed worse results than only considering the positives.

\smallskip\noindent\textbf{Multiple task loss.}
For our experiments, we augment the architectures of SiamFC~\cite{bertinetto2016fully} and SiamRPN~\cite{li2018high} with our segmentation branch and the loss $\mathcal{L}_{mask}$, obtaining what we call the \emph{two-branch} and \emph{three-branch} variants of SiamMask.
These respectively optimise the multi-task losses $\mathcal{L}_{2B}$ and $\mathcal{L}_{3B}$, defined as:
\begin{equation}
\label{eq:2b}
\mathcal{L}_{2B} = \lambda_{1} \cdot \mathcal{L}_{mask} + \lambda_{2} \cdot \mathcal{L}_{sim},
\end{equation}
\begin{equation}
\label{eq:3b}
\mathcal{L}_{3B} = \lambda_{1} \cdot \mathcal{L}_{mask} + \lambda_{2} \cdot \mathcal{L}_{score}+ \lambda_{3} \cdot \mathcal{L}_{reg}.
\end{equation}
We did not search over the hyperparameters of Eq.~\ref{eq:2b} and Eq.~\ref{eq:3b} and simply set $\lambda_1=32$ like in~\cite{o2015learning} and $\lambda_{2}=\lambda_{3}=1$.
The task-specific branches for the box and score outputs are constituted by two $1{\times}1$ convolutional layers.

\subsection{Box generation}
\label{sec:box_generation}
\begin{figure}[t]
	\centering
	\includegraphics[width=\columnwidth]{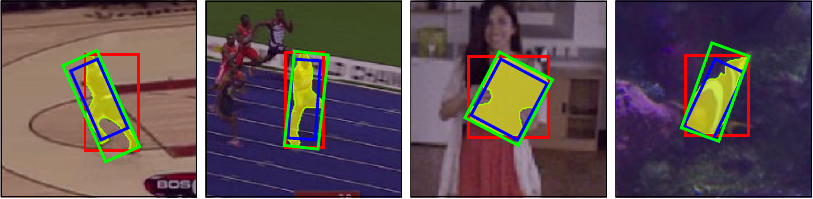}
	\caption{In order to generate a bounding box from a binary mask (in yellow), we experiment with three different methods.
\textit{Min-max}: the axis-aligned rectangle containing the object (red); \textit{MBR}: the minimum bounding rectangle (green); \textit{Opt}: the rectangle obtained via the optimisation strategy proposed in VOT-2016~\cite{kristan2016novel,kristan2018sixth} (blue).}
\label{fig:object_representations}
\end{figure}

Note that, while video object segmentation benchmarks require binary masks, typical tracking benchmarks such as VOT~\cite{kristan2016novel} require a bounding box as final representation of the target object.
We consider three different strategies to generate a bounding box from a binary mask, represented in Fig.~\ref{fig:object_representations}:
(1) axis-aligned bounding rectangle (\emph{Min-max}), (2) rotated minimum bounding rectangle (\emph{MBR}) and (3) the optimisation strategy used for the automatic bounding box generation proposed in the VOT benchmarks~\cite{kristan2016novel,kristan2018sixth} (\emph{Opt}).
We empirically evaluate these alternatives in Section~\ref{sec:experiments} (Table~\ref{tab:iou}).

\subsection{Training and testing}
\label{sec:training_testing}
During training, pairs of exemplar and search image samples are input into the network, and the predicted masks, scores, and boxes are used to optimize the multi-task loss (Eq.~\ref{eq:2b} or~\ref{eq:3b}).
Positive and negative samples are defined differently in the two- and three-branch version of SiamMask.
For $\mathcal{L}_{3B}$, a RoW is considered positive ($y_{n} = 1$) if one of its anchor boxes has an intersection-over-union (IOU) with the ground-truth bounding-box of at least 0.6, and negative ($y_{n} = -1$) otherwise.
For $\mathcal{L}_{2B}$, we adopt a similar strategy to the one of~\cite{bertinetto2016fully} to define positive and negative samples: a RoW is considered a positive sample if the distance between the center of the prediction and the center of the ground-truth is below 16 pixels (in feature space), and negative otherwise.

During online tracking, the offline-trained SiamMask is used to produce masks and boxes for every input frame, with no further adaptation of the network's parameters and a simple axis-aligned bounding-box initialization.
In both variants, the output mask is selected using the location that achieves the maximum score in the classification branch.

% \section{Segmentation-Based Multiple Object Tracking}
\section{Multiple Object Tracking and Segmentation}
\label{sec:MOT}
\begin{figure*}[t]
	\centering
	\includegraphics[width=\textwidth]{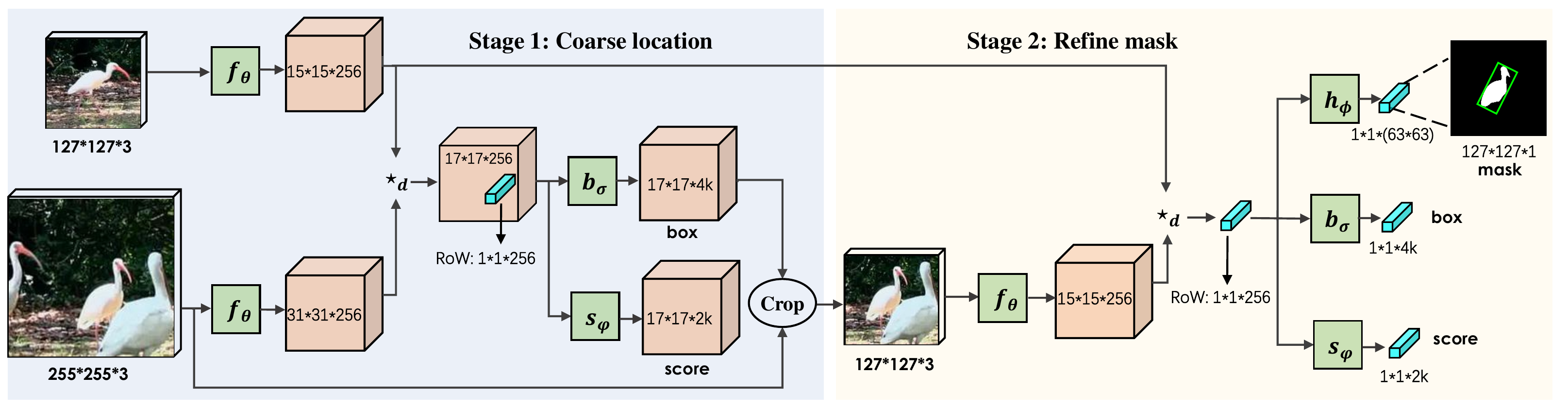}
	\caption{The two-stage version of SiamMask used for each object in the multiple object tracking and segmentation problem.}
	\label{fig:cascaded}
\end{figure*}

We extend the application of the proposed SiamMask to segmentation-based multiple object tracking~\cite{wang2019empirical}, which requires to segment and track an arbitrary number of objects throughout the video~\cite{zhang2008global}.
Compared to the single-object tracking problem, it presents the added difficulty of requiring to disambiguate between object instances.
This means that, across frames, each object needs to be labelled with the correct identity, which becomes particularly challenging in crowded scenes.

To address this problem, we use a pre-trained segmentation-based object detection algorithm to initialize individual tracks, and then for each object we apply SiamMask twice in a ``cascaded fashion''.
More specifically, for each new frame an off-the-shelf segmentation-based object detector~\cite{chen2019hybrid} is used to obtain $M$ candidate masks $\mathcal{D}=\{d_i\}_{i=1}^M$.
Given existing trajectories $\mathcal{T}=\{t_i\}_{i=1}^N$, SiamMask is used to produce $N$ masks $\mathcal{P}=\{p_j\}_{j=1}^N$.
%of the objects corresponding to the trajectories in $\mathcal{T}$.
Then, identity association across frames is obtained by solving the assignment between $\mathcal{D}$ and $\mathcal{P}$ as an optimal transport problem.
Let $a_{ij}$ be the pairwise affinity between the $i$-th mask $d_i$ in $\mathcal{D}$ and the $j$-th mask $p_j$ in $\mathcal{P}$.
It is defined as the IOU (intersection over union) between the masks $d_i$ and $p_j$.
Let $x_{ij} \in \{0,1\}$ represent the pairwise association between $d_i$ and $p_j$, which is 1 if $d_i$ and $p_j$ belong to the same object, and 0 otherwise.
The assignment between $\mathcal{D}$ and $\mathcal{P}$ is formulated as:
\begin{equation}\label{eq:assignment}
    \mathop{max}\limits_{\{x_{ij}\}}\sum_{i=1}^M \sum_{j=1}^N a_{ij}x_{ij},
\end{equation}

\begin{equation*}
    s.t. \left\{
    \begin{array}{ll}
        \forall i \in \{1,2,...,M\} & \sum_{j=1}^N x_{ij} \leq 1, \\
        \forall i \in \{1,2,...,N\} & \sum_{i=1}^M x_{ij} \leq 1.
    \end{array}\right.
\end{equation*}
This ensures that each mask in $\mathcal{D}$ is associates with at most one mask in $\mathcal{P}$, and conversely that each mask in $\mathcal{P}$ is associated with at most one mask in $\mathcal{D}$.
This constrained integer optimization problem can be solved by the Hungarian algorithm.
By solving this assignment problem at each frame, objects tracks are maintained throughout the video.

Although many existing visual object tracking algorithms and object segmentation algorithms can be used to construct object correspondences between adjacent frames, we found SiamMask to be a practical and effective choice.
Its low computational cost allows us to deal with multiple objects, and
the tracking score from the classification branch can be conveniently used to indicate the occlusion or disappearance of the target object.

For our experiments on multiple object tracking and segmentation we propose a cascaded version of SiamMask, as illustrated in Fig.~\ref{fig:cascaded}.
In a first stage, the regression branch of SiamMask predicts a coarse location of each object.
In the second, a crop from the search image is extracted in correspondence of the bounding box corresponding to the highest score in the first stage, and used to predict a refined mask from the segmentation branch.
These masks are the new predictions $\mathcal{P}$ associated with the current object trajectories.
Then, IoUs between the masks predicted by SiamMask and the newly detected masks $\mathcal{D}$ are computed.
The assignment from Eq.~\ref{eq:assignment} associates the detected masks $\mathcal{D}$ with the existing trajectories, and it is also used to keep track of newly appeared or disappeared objects ($|M-N|$ is the number of new or lost objects). 

\section{Experiments}
\label{sec:experiments}
In this experimental section, we first describe implementation details (\ref{sec:implementation}) and then we evaluate our approach on three related but different problems: visual object tracking on VOT-2016, VOT-2018, GOT-10k and TrackingNet (~\ref{sec:tracking_evaluation}); video object segmentation on DAVIS-2016, DAVIS-2017 and YouTube-VOS (~\ref{sec:vos_evaluation}); and multiple-object tracking and segmentation on YouTube-VIS (~\ref{sec:mot_evaluation}).
We conclude the section with ablation studies (~\ref{sec:ablations}) and qualitative examples from benchmarks videos (~\ref{sec:qualitative_results}).

\subsection{Implementation}
\label{sec:implementation}
\smallskip\noindent\textbf{Network architecture.}
For both the two-branch and three-branch variants of SiamMask, we use a ResNet-50~\cite{he2016deep} architecture until the final convolutional layer of the \mbox{$4$-th} stage as our backbone $f_\theta$.
In order to obtain a high spatial resolution in deeper layers, we reduce the output stride to $8$ (from $32$) by using convolutions with stride 1.
Moreover, we increase the receptive field by using dilated convolutions~\cite{chen2017deeplab}.
Table~\ref{tab:backbone_architecture} outlines the structure of $f_{\theta}$, while Table~\ref{two_branches} and~\ref{three_branches} show the architectures of the branches of the two variants of SiamMask.
In our model, we add to the shared backbone $f_{\theta}$ a (not shared) ``\emph{adjust} layer'' constituting of a $1{\times}1$ \textit{conv} with $256$ outputs (for simplicity, this is omitted Eq.~\ref{eq:cross}).
In the $3{\times}3$ convolutional layer of block conv4\_1, the stride is 1 and the dilation rate is 2.
The conv5 block in both variants contains a normalization layer and a ReLU non-linearity, and block conv6 only consists of a ${1\times}1$ convolutional layer.
For the three-branch variant, the number $k$ of anchors is set to 5.
Exemplar and search images share the network parameters from block conv1 to conv4\_x, and do not share the parameters of the adjust layer.
As a last step, depth-wise cross-correlation is carried between the output feature maps from the individual adjust layers for the exemplar and search area, obtaining an output of size $17{\times}17{\times}256$.

\begin{table}
\caption{Backbone architecture. The structure of each block is shown in square brackets}
\resizebox{\columnwidth}{!}
{
\begin{tabular}{|c|c|c|c|}
\hline Block & Exemplar output size & Search output size & Details \\
\hline conv1\_x & $61{\times}61$ & $125{\times}125$ & $7{\times}7$, 64, stride 2 \\
\hline & & & $3{\times}3$ max pool, stride 2 \\
conv2\_x & $31{\times}31$ & $63{\times}63$ &
    $\left[\begin{array}{cc}1{\times}1, & 64 \\
    3\times 3, & 64 \\
    1{\times}1, & 256\end{array}\right]{\times}3$ \\
\hline conv3 x & $15{\times}15$ & $31{\times}31$ &
    $\left[\begin{array}{cc}1{\times}1, & 128 \\
    3{\times}3, & 128 \\
    1{\times}1, & 512\end{array}\right]{\times}4$ \\
\hline conv4 x & $15{\times}15$ & $31{\times}31$ &
    $\left[\begin{array}{cc}1{\times}1, & 256 \\
    3{\times}3, & 256 \\
    1{\times}1, & 1024\end{array}\right]{\times}6$ \\
\hline adjust & $15{\times}15$ & $31{\times}31$ & $1{\times}1,256$ \\
\hline xcorr & \multicolumn{2}{|c|}{$17{\times}17$} & depth-wise \\
\hline
\end{tabular}
}
\label{tab:backbone_architecture}
\end{table}

\begin{table}[t]
    \caption{The architecture of the two-branch variant.}
    \centering
    \begin{tabular}{|c|c|c|}
    \hline Block & Score & Mask \\
    \hline conv5 & $1{\times}1,256$ & $1{\times}1,256$ \\
    \hline conv6 & $1{\times}1,1$ & $1{\times}1,(63{\times}63)$ \\
    \hline
    \end{tabular}
\label{two_branches}
\end{table}

\begin{table}[t]
    \caption{The architecture of the three-branch variant. $k$ is the number of anchors for each RoW.}
    \centering
    \begin{tabular}{|c|c|c|c|}
    \hline Block & Score & Box & Mask \\
    \hline conv5 & $1{\times}1,256$ & $1{\times}1,256$ & $1{\times}1,256$ \\
    \hline conv6 & $1{\times}1,2 k$ & $1{\times}1,4 k$ & $1{\times}1,(63{\times}63)$ \\
    \hline
\end{tabular}
\label{three_branches}
\end{table}

\begin{table}[t]
\caption{Accuracies for different bounding box representation strategies on VOT-2016.}
\resizebox{\columnwidth}{!}{
\begin{tabular}{|c|c|c|c|c|}
\hline \multicolumn{2}{|c|}{ Method } & mIoU $(\%)$ & mAP@0.5 IoU & mAP@0.7 IoU \\
\hline \multirow{2}{*}{ Oracle } & Fixed & $73.43$ & $90.15$ & $62.52$ \\
\cline { 2 - 5 } & Min-max/aligned & $77.70$ & $88.84$ & $65.16$ \\
\cline { 2 - 5 } & MBR & $84.07$ & $97.77$ & $80.68$ \\
\hline \multicolumn{2}{|c|}{ SiamFC \cite{bertinetto2016fully} } & $50.48$ & $56.42$ & $9.28$ \\
\hline \multicolumn{2}{|c|}{ SiamRPN \cite{li2018high} } & $60.02$ & $76.20$ & $32.47$ \\
\hline \multirow{3}{*}{ SiamMask } & Min-max/aligned & $65.05$ & $82.99$ & $43.09$ \\
\cline { 2 - 5 } & MBR & $67.15$ & $85.42$ & $50.86$ \\
\cline { 2 - 5 } & Opt & $\mathbf{7 1 . 6 8}$ & $\mathbf{9 0 . 7 7}$ & $\mathbf{6 0 . 4 7}$ \\
\hline
\end{tabular}}
\label{tab:iou}
\end{table}

\smallskip\noindent\textbf{Offline training settings.}
As in SiamFC \cite{bertinetto2016fully}, the size of exemplar and search image patches are $127{\times}127$ and $255{\times}255$ pixels respectively.
Training samples originate from COCO \cite{lin2014microsoft}, ImageNet-VID \cite{russakovsky2015imagenet} and YouTube-VOS \cite{xu2018youtube}, and were augmented by randomly shifting and scaling the input patches.
Random translations are within $\pm 4$ pixels for exemplar images and within $\pm 64$ pixels for search images.
Random scaling is within [0.95, 1.05] and [0.82, 1.18] for exemplar and search images respectively.
The network backbone is pre-trained on the ILSVRC ImageNet classification task (1000 classes).
We use SGD with a first \emph{warmup} phase in which the learning rate increases linearly from $10^{-3}$ to $5{\times}10^{-3}$ for the first 5 epochs and then descreases logarithmically until $5{\times}10^{-4}$ for 15 more epochs.

\smallskip\noindent\textbf{Online inference settings.}
During tracking, SiamMask is simply evaluated once per frame, without any adaptation.
In both our variants, we select the output mask using the location attaining the maximum score in the classification branch.
Then, after having applied a per-pixel sigmoid, we binarise the output of the mask branch at the threshold of $0.5$.
In the \textit{two-branch} variant, for each video frame after the first one, we fit the output mask with the \emph{Min-max} box and use it as reference to crop the next frame search region.
Instead, in the \textit{three-branch} variant, we find more effective to exploit the highest-scoring output of the box branch as reference.

\smallskip\noindent\textbf{Timing.}
SiamMask operates online without any adaptation to the test sequence.
On a single NVIDIA RTX 2080 GPU, we measured an average speed of 55 and 60 frames per second, respectively for the \emph{two-branch} and \emph{three-branch} variants.
Note that SiamMask does not perform online adaptation of the network parameters during tracking, and that the highest computational burden comes from the feature extractor $f_{\theta}$.

\smallskip\noindent\textbf{Datasets.}
To evaluate tracking performance, the following four benchmarks were used: VOT-2016 \cite{kristan2017visual}, VOT-2018 \cite{kristan2018sixth}, GOT-10k \cite{huang2019got}, and TrackingNet \cite{muller2018trackingnet}.
\begin{itemize}
    \item \textbf{VOT-2016 and VOT-2018.}
    We use VOT-2016 to understand how different types of representation affect the performance.
    For this first experiment, we use mean intersection over union (IOU) and Average Precision (AP)@$\{0.5,0.7\}$ IOU.
    We then compare against the state-of-the-art on both VOT-2018 and VOT-2016, using the official VOT toolkit and the Expected Average Overlap (EAO), a measure that considers both accuracy and robustness of a tracker~\cite{kristan2016novel}.
    \item \textbf{GOT-10k and TrackingNet.} These are larger and more recent visual object tracking datasets which are useful to test the generalization ability of trackers on a vast number of diverse classes, scenarios, and types of motions.
\end{itemize}

To evaluate (VOS) segmentation performance, the following four benchmarks were used: DAVIS-2016 \cite{perazzi2016benchmark}, DAVIS-2017 \cite{pont20172017}, YouTube-VOS \cite{xu2018youtube}, and YouTube-VIS \cite{yang2019video}.
\begin{itemize}
    \item \textbf{DAVIS-2016 and DAVIS-2017.}
    We report the performance of SiamMask on DAVIS-2016~\cite{perazzi2016benchmark}, DAVIS-2017~\cite{pont20172017} and YouTube-VOS~\cite{xu2018youtube} benchmarks.
    For both DAVIS datasets, we use the official performance measures: the Jaccard index ($\mathcal{J}$) to express region similarity and the F-measure ($\mathcal{F}$) to express contour accuracy.
    For each measure $\mathcal{C} \in \{\mathcal{J}, \mathcal{F}\}$, three statistics are considered: mean $\mathcal{C}_{\mathcal{M}}$, recall $\mathcal{C}_{\mathcal{O}}$, and decay $\mathcal{C}_{\mathcal{D}}$, which informs us about the gain/loss of performance over time~\cite{perazzi2016benchmark}.
    \item \textbf{YouTube-VOS.}
    Following \cite{xu2018youtube}, the final result for a test sample in YouTube-VOS is the average of the following four metrics: the overlap precision $\mathcal{J_S}$ for seen classes that appear in the training set, the overlap precision $\mathcal{J_U}$ for unseen classes that do not appear in the training set, the edge precision $\mathcal{F_S}$ for seen classes, and the edge precision $\mathcal{F_U}$ for unseen classes.
    We report the mean Jaccard index and F-measure for both seen ($\mathcal{J}_{\mathcal{S}}$, $\mathcal{F}_{\mathcal{S}}$) and unseen categories ($\mathcal{J}_{\mathcal{U}}$, $\mathcal{F}_{\mathcal{U}}$).
    $\mathcal{O}$ is the average of these four measures.
    \item \textbf{YouTube-VIS.}
    This is a large multiple object tracking and segmentation dataset
    with 2,883 high-resolution videos containing objects labelled with 40 classes, and 131,000 high-quality pixel-wise masks.
    Average precision (AP) and average recall are used as performance metrics \cite{yang2019video}.
\end{itemize}

\subsection{Evaluation for tracking}
\label{sec:tracking_evaluation}
\noindent\textbf{Target object representation.}
Existing tracking methods typically predict axis-aligned bounding boxes with a fixed~\cite{bertinetto2016fully,henriques2014high,danelljan2015learning,lukezic2017discriminative} or variable~\cite{li2018high,held2016learning,zhu2018distractor} aspect ratio.
We are interested in understanding to which extent producing a per-frame binary mask can improve tracking.
In order to focus on representation accuracy, for this experiment only we ignore the temporal aspect and sample video frames at random.
The approaches described in the following paragraph are tested on randomly cropped search patches (with random shifts within $\pm16$ pixels and scale deformations up to $2^{1\pm0.25}$) from the sequences of VOT-2016.

In Table~\ref{tab:iou}, we compare our \emph{three-branch} variant using the \emph{Min-max}, \emph{MBR} and \emph{Opt} approaches (described in Section~\ref{sec:box_generation} and in Figure~\ref{fig:object_representations}).
For reference, we also report results for SiamFC and SiamRPN as representative of the fixed and variable aspect-ratio approaches, together with three \emph{oracles} that have access to per-frame ground-truth information and serve as upper bounds for the different representation strategies.
(1) The fixed aspect-ratio oracle (``fixed" in the table) uses the per-frame ground-truth area and center location, but fixes the aspect reatio to the one of the first frame and produces an axis-aligned bounding box.
(2) The \emph{Min-max} oracle uses the minimal enclosing rectangle of the rotated ground-truth bounding box to produce an axis-aligned bounding box.
(3) Finally, the \textit{MBR} oracle uses the rotated minimum bounding rectangle of the ground-truth.
Note that (1), (2) and (3) can be considered, respectively, the performance upper bounds for the representation strategies of SiamFC, SiamRPN and SiamMask.

The results are reported for SiamFC and SiamRPN as the representative trackers using, respectively, fixed and variable aspect-ratio bounding-box representation.
We use SiamMask's three-branch variant, for which report the results obtained when using Min-max, MBR, and Opt representation strategies.
Although SiamMask-\textit{Opt} offers the highest IOU and mAP, it requires significant computational resources due to its slow optimisation procedure.
SiamMask-\textit{MBR} achieves a mAP@0.5 IOU  of $85.4$, with a respective improvement of $+29$ and $+9.2$ points w.r.t. the two fully-convolutional baselines.
Interestingly, the gap significantly widens when considering mAP at the higher accuracy regime of 0.7 IOU: $+41.6$ and $+18.4$ respectively.
Notably, our accuracy results are not far from the fixed aspect-ratio oracle.
Moreover, comparing the upper bound performance represented by the oracles, it is possible to notice how, by simply changing the bounding box representation, there is a great room for improvement (\emph{e.g.} $+10.6\%$ mIOU improvement between the fixed aspect-ratio and the \textit{MBR} oracles).

Overall, this study shows how the \textit{MBR} strategy to obtain a rotated bounding box from a binary mask of the object offers a significant advantage over popular strategies that simply report axis-aligned bounding boxes.

\smallskip\noindent\textbf{Results on VOT-2016 and VOT-2018.}
Table~\ref{tab:vot18and16} shows the results of SiamMask with different box generation strategies on the VOT-2016 and VOT-2018 benchmarks.
The metrics EAO, accuracy, robustness, and speed were considered.
SiamMask-box indicates that the box branch of SiamMask is adopted for inference despite the mask branch has been trained.
From the table, the following observations can be made:
\begin{itemize}
    \item On the simpler VOT-2016 benchmark, compared with SiamMask-box which directly outputs axis-aligned boxes from the box regression branch, SiamMask-Opt which outputs boxes from the mask branch increases EAO by 3\%, and increases the accuracy by 4.7\%.
    \item On the more challenging VOT-2018, SiamMask-Opt increases EAO by 2.4\% and increases the accuracy by 5.8\%. The robustness is also increased.
    \item Overall, SiamMask-MBR yields a better performance than SiamMask-box, while keeping real time speed.
\end{itemize}
In general, we can observe clear improvements on all evaluation metrics by using the mask branch for box generation.
Note how SiamMask-Opt is best for overall EAO (especially in terms of Accuracy), but its improvement w.r.t. SiamMask-MBR does not justify the significantly higher computational cost.

\begin{table}[t]
\caption{Results of SiamMask on the VOT-2016 and VOT-2018 benchmarks.}
\resizebox{\columnwidth}{!}{
\begin{tabular}{|c|c|c|c|c|c|c|c|}
\hline \multirow{2}{*}{ Method } & \multicolumn{3}{|c|}{ VOT-2016 benchmark } & \multicolumn{3}{c|}{ VOT-2018 benchmark } & Speed \\
\cline { 2 - 7 } & EAO & Accu. $\uparrow$ & Robust. $\downarrow$ & EAO $\uparrow$ & Accu. $\uparrow$ & Robust. $\downarrow$ & (fps) $\uparrow$ \\
\hline SiamMask-box & $0.412$ & $0.623$ & $0.233$ & $0.363$ & $0.584$ & $0.300$ & $\mathbf{7 6}$ \\
\hline SiamMask-MBR & $0.433$ & $0.639$ & $\mathbf{0 . 2 1 4}$ & $0.380$ & $0.609$ & $\mathbf{0 . 2 7 6}$ & 55 \\
\hline SiamMask-Opt & $\mathbf{0 . 4 4 2}$ & $\mathbf{0 . 6 7 0}$ & $0.233$ & $\mathbf{0 . 3 8 7}$ & $\mathbf{0 . 6 4 2}$ & $0.295$ & 5 \\
\hline
\end{tabular}}
\label{tab:vot18and16}
\end{table}

In Table~\ref{tab:vot2018} we compare the two variants of SiamMask with \textit{MBR} strategy and SiamMask--\textit{Opt} against five popular trackers on the VOT-2018 benchmark.
Unless stated otherwise, SiamMask refers to our \textit{three-branch} variant with \textit{MBR} strategy.
Both variants achieve a strong performance and run in real-time.
In particular, our \textit{three-branch} variant significantly outperforms DaSiamRPN~\cite{zhu2018distractor} (which is trained on YouTube-bounding-boxes~\cite{real2017youtube}), achieving a EAO of $0.380$ while running at 55 frames per second.
Even without box regression branch, our simpler \textit{two-branch} variant (SiamMask-2B) achieves a high EAO of $0.334$, which is in par with SA\_Siam\_R~\cite{he2018towards} and superior to any other real-time method in the published literature at the time of the conference version of this paper~\cite{wang2019fast}.
Finally, in SiamMask--\textit{Opt}, the strategy proposed in~\cite{kristan2018sixth} to find the optimal rotated rectangle from a binary mask brings the best overall performance (and a particularly high accuracy), but comes at a significant computational cost.

Our model is particularly strong under the accuracy metric, showing a significant advantage with respect to the Correlation Filter-based trackers CSRDCF~\cite{lukezic2017discriminative}, STRCF~\cite{li2018learning}.
This is not surprising, as SiamMask relies on a richer object representation, as demonstrated in the experiments of Table~\ref{tab:iou}.
Interestingly, similarly to us, He \emph{et al.} (SA\_Siam\_R)~\cite{he2018towards} are motivated to achieve a more accurate target representation by considering multiple rotated and rescaled bounding boxes.
However, their representation is still constrained to a fixed aspect-ratio box.

\begin{table}[t]
\caption{Comparison with the state-of-the-art on the VOT-2018 benchmark.}
\resizebox{\columnwidth}{!}{
\begin{tabular}{|c|c|c|c|c|c|c|c|c|}
\hline
\multirow{3}{*}{Metric} & \multicolumn{3}{c|}{SiamMask} & \multirow{3}{*}{DaSiam-RPN \cite{zhu2018distractor}} & \multirow{3}{*}{Siam-RPN \cite{li2018high}} & \multirow{3}{*}{SA\_Siam\_R \cite{henriques2014high}} & \multirow{3}{*}{CSRDCF \cite{lukezic2017discriminative}} & \multirow{3}{*}{STRCF \cite{li2018learning}} \\
\cline{2-4}
 & \multirow{2}{*}{Opt} & \multicolumn{2}{c|}{MBR} & & & & & \\
 \cline{3-4}
  & & Three-branch & Two-branch & & & & & \\
  \hline EAO $\uparrow$ & $\mathbf{0 . 3 8 7}$ & $\mathbf{0 . 3 8 0}$ & $0.334$ & $0.326$ & $0.244$ & $0.337$ & $0.263$ & $0.345$ \\
\hline Accur. $\uparrow$ & $\mathbf{0 . 6 4 2}$ & $\mathbf{0 . 6 0 9}$ & $0.575$ & $0.569$ & $0.490$ & $0.566$ & $0.466$ & $0.523$ \\
\hline Robus. $\downarrow$ & $0.295$ & $0.276$ & $0.304$ & $0.337$ & $0.460$ & $0.258$ & $0.318$ & $\mathbf{0 . 2 1 5}$ \\
\hline Speed $\uparrow$ & 5 & 55 & 60 & 160 & $\mathbf{2 0 0}$ & $32.4$ & $48.9$ & $2.9$ \\
\hline
\end{tabular}}
\label{tab:vot2018}
\end{table}

\begin{figure}[t]
	\centering
	\includegraphics[width=\columnwidth]{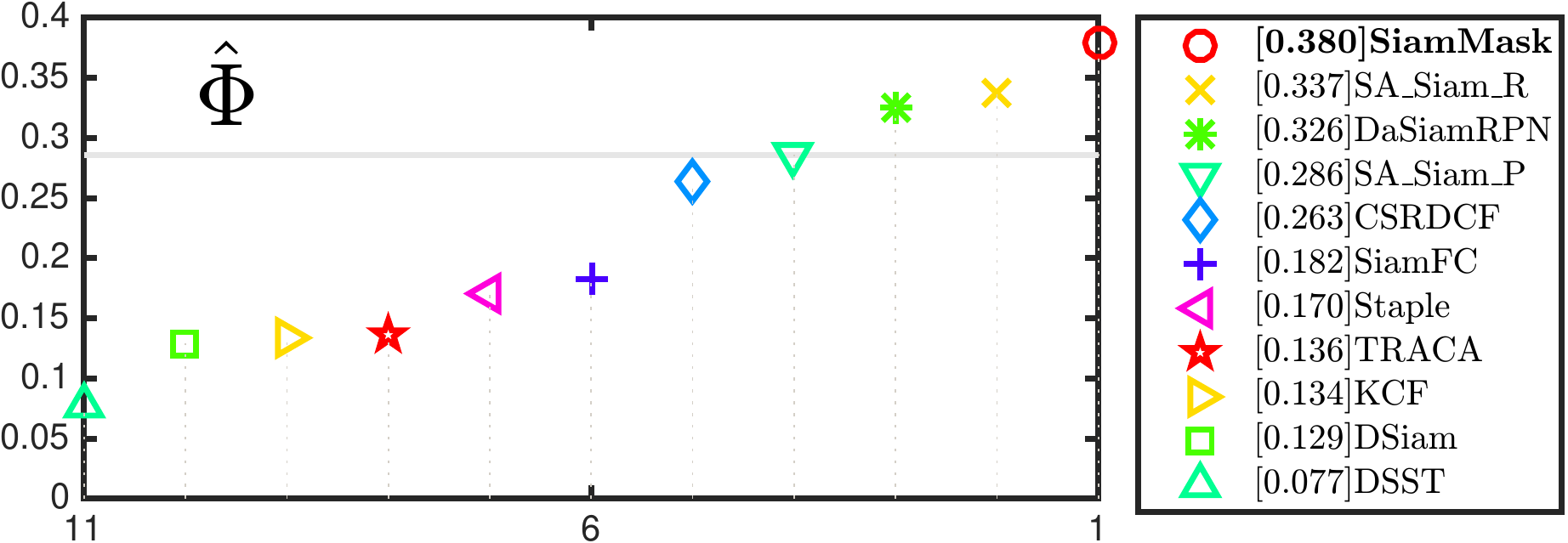}
	\caption{The EAO plot for the proposed SiamMask and the top 10 real time competing trackers on the VOT2018 challenge.}
\label{fig:realtime_comparison}
\end{figure}

\smallskip\noindent\textbf{Real-time VOT-2018 comparison.}
As additional visualization, we use the plots provided by VOT toolkit to compare SiaMask against the top 10 real-time trackers in terms of EAO (Expected Average Overlap: the metric used as a summary by VOT).
In Fig.~\ref{fig:realtime_comparison}, the horizontal coordinate represents the rank of the trackers, and the vertical coordinate their EAO.
The horizontal gray line in the figure is what is considered by the VOT committee as the state-of-the-art at the time of the competition.
Compared with the fully convolutional network SiamFC, SiamMask increases the performance by a very significant \textit{absolute} 19.8\%.

\begin{figure}[t]
	\centering
	\subfigure[]{
	\includegraphics[width=\columnwidth]{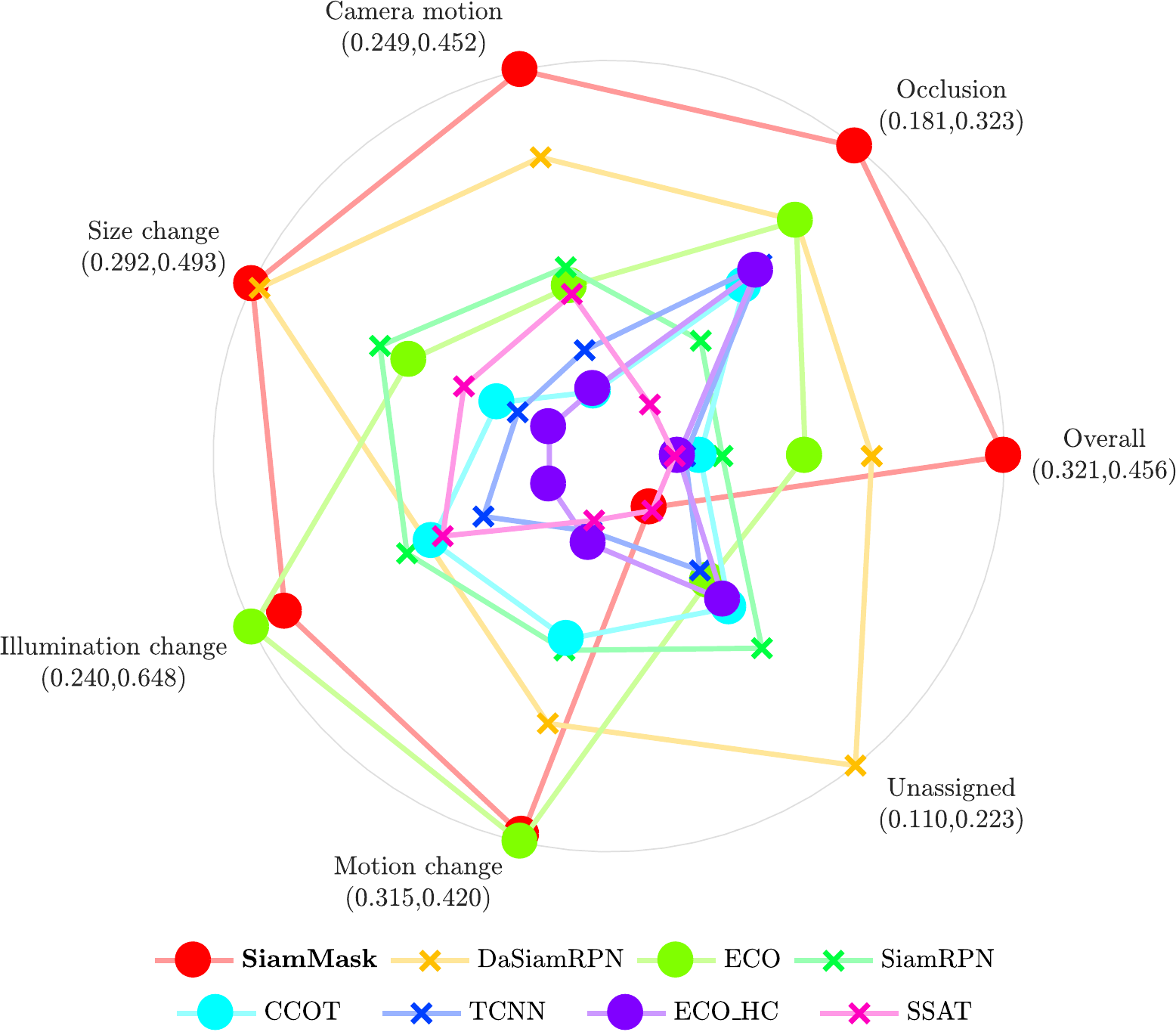}
	}
	\subfigure[]{\includegraphics[width=\columnwidth]{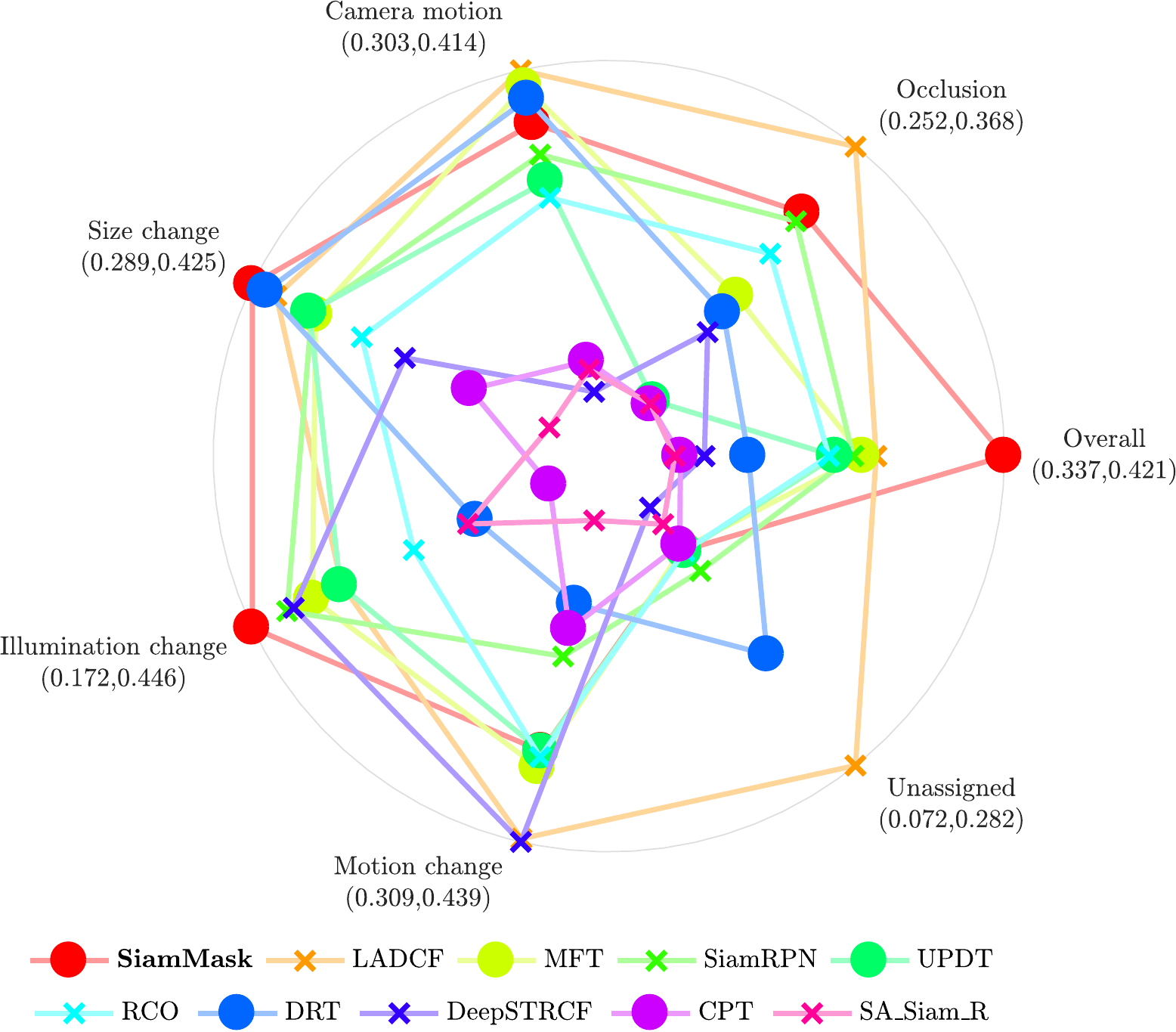}}
	\caption{Comparison between SiamMask and state-of-the-art trackers with respect to different visual scene attributes on VOT2016 and VOT2018.}
\label{fig:scene_attributes}
\end{figure}

\smallskip\noindent\textbf{VOT attributes breakdown.}
In the VOT benchmarks, frames are densely labelled with scene attributes to give a more qualitative understanding of how different trackers perform under different circumstances.
The scene attributes are: occlusion, illumination change, motion change, size (scale) change, and camera motion.
We compared SiamMask with popular and representative trackers~\cite{danelljan2017eco,he2018towards,li2018high,li2018learning,zhu2018distractor,danelljan2016beyond,bhat2018unveiling,nam2016modeling} with respect to these attributes on the VOT-2016 and VOT-2018 benchmarks.
The results are shown in Fig.~\ref{fig:scene_attributes}.
For both benchmarks, it can be seen that SiamMask obtains the best results for most scene attributes.
One clear advantage brought by our method is the capability of providing pixel-wise mask representation for the target object (at a high speed), which allows a much higher accuracy and ease of adaptation, especially in presence of rapid non-rigid deformations.

\smallskip\noindent\textbf{Unsupervised learning.}
A recent trend in object tracking (and computer vision in general) is to train feature extractors on large scale datasets with a self-supervised proxy task.
This is a compelling strategy, as it offers a way to exploit large datasets without having to provide costly box or mask labels.
However, these methods encounter the additional challenge of choosing the right proxy task outside of the usual supervision loop, which requires a much larger amount of experiments for tuning a large number of highly-consequential hyperparameters, like the ones controlling data augmentation.
Table~\ref{tab:SSL_trackers} compares our method with a representative set of recent self-supervised-learning-based trackers~\cite{wang2021unsupervised,yuan2020self,zheng2021learning,sio2020s2siamfc,wu2021progressive} on the VOT-2018 benchmark.
It can be seen that, while their speed is comparable to our method (and sometimes faster), their overall performance still lags behind.

\begin{table}[t]
\caption{Comparison with unsupervised learning-based methods on the VOT-2018 benchmark.}
\resizebox{\columnwidth}{!}{
\begin{tabular}{|c|c|c|c|c|c|c|}
\hline Metric & LUDT \cite{wang2021unsupervised} & CycleSiam \cite{yuan2020self} & USOT \cite{zheng2021learning} & S2SiamFC \cite{sio2020s2siamfc} & PUL \cite{wu2021progressive} & Our method \\
\hline EAO $\uparrow$ & $0.230$ & $0.317$ & $0.344$ & $0.180$ & $0.203$ & $\mathbf{0 . 3 8 0}$ \\
\hline Accuracy $\uparrow$ & $0.49$ & $0.549$ & $0.578$ & $0.463$ & $0.516$ & $\mathbf{0 . 6 0 9}$ \\
\hline Robustness $\downarrow$ & $0.412$ & $0.314$ & $0.304$ & $0.782$ & $0.660$ & $\mathbf{0 . 2 7 6}$ \\
\hline Speed (fps) $\uparrow$ & 55 & 44 & $-$ & $\mathbf{86}$ & $-$ & $55$ \\
\hline
\end{tabular}}
\label{tab:SSL_trackers}
\end{table}

\smallskip\noindent\textbf{Results on GOT-10k and TrackingNet.}
GOT-10k \cite{huang2019got} is a very large scale tracking dataset covering 563 object classes and 87 motion patterns.
In total, it consists of 10,000 video clips with 1.5 million bounding-box labels.
Trackers are evaluated on a selection of 180 videos exhibiting 84 different object classes and 32 ``motion types''.
Trackers are ranked using average overlap.
We also reported the success rates at two thresholds: 0.5 and 0.75.
In this case, SiamMask was compared with the results made available by the benchmark for CFNet \cite{valmadre2017end}, SiamFC \cite{bertinetto2016fully}, GOTURN \cite{he2016deep}, CCOT \cite{danelljan2016beyond} and MDNet \cite{nam2016learning}.
Results are shown in Table~\ref{tab:got10k}.
Compared to CFNet \cite{valmadre2017end} (which has best performance among the competing trackers), SiamMask has a significant advantage on all the metrics considered.
In general, the fact that SiamMask maintains its strong performance on a dataset with a large number of classes should be taken as a positive signal for its generalization capabilities.
It provides a relative increase in average overlap of 37\%, and of up to 150\% in success rate.
However, it is hard to establish an apple-to-apple comparison between SiamMask and the methods reported by this benchmark.
On the one hand, the trackers reported by the benchmark are trained on the training split of the same dataset (which is supposedly sampled from the same distribution as the set used as benchmark).
On the other, a part from the ``person'' class, the GOT-10k training set does not contain any other class from the benchmark set.
For the sake of simplicity, and for consistency with the other experiments in the paper, we did not enforce the same separation, so we do not have data regarding the class overlap between the two sets.
This should be taken into account when doing comparisons on this benchmark.

\begin{table}
\caption{Comparison on GOT-10k benchmark.}
\resizebox{\columnwidth}{!}{
\begin{tabular}{|c|c|c|c|c|c|c|}
\hline Metric & SiamMask & CFNet \cite{valmadre2017end} & SiamFC \cite{bertinetto2016fully} & GOTURN \cite{he2016deep} & CCOT \cite{danelljan2016beyond} & MDNet \cite{nam2016learning} \\
\hline Average overlap & $51.4$ & $37.4$ & $34.8$ & $34.2$ & $32.5$ & $29.9$ \\
\hline Success rate with overlap threshold $=0.75$ & $36.6$ & $14.4$ & $9.8$ & $12.4$ & $10.7$ & $9.9$ \\
\hline Success rate with overlap threshold $=0.5$ & $58.7$ & $40.4$ & $35.3$ & $37.5$ & $32.8$ & $30.3$ \\
\hline
\end{tabular}}
\label{tab:got10k}
\end{table}

\begin{table}[t]
\caption{Comparison on TrackingNet benchmark.}
\resizebox{\columnwidth}{!}{
\begin{tabular}{|c|c|c|c|c|c|c|}
\hline Metric & SiamMask & ATOM \cite{danelljan2019atom} & MDNet \cite{nam2016learning} & CFNet \cite{valmadre2017end} & SiamFC \cite{bertinetto2016fully} & ECO \cite{danelljan2017eco} \\
\hline AUC of success rate & $\mathbf{7 2 . 5}$ & $70.3$ & $60.6$ & $57.8$ & $57.1$ & $55.4$ \\
\hline Tracking precision & $\mathbf{6 6 . 4}$ & $64.8$ & $56.5$ & $53.3$ & $53.3$ & $49.2$ \\
\hline Normalized precision & $\mathbf{7 7 . 8}$ & $77.1$ & $70.5$ & $65.4$ & $66.3$ & $61.8$ \\
\hline
\end{tabular}}
\label{tab:trackingnet}
\end{table}

TrackingNet \cite{muller2018trackingnet} is a popular large video benchmark of 511 videos for testing visual object tracking algorithms.
Trackers were ranked according to the area under the curve (AUC) of the success rate, tracking precision, and normalized precision.
On this dataset, SiamMask was compared with ATOM \cite{danelljan2019atom}, MDNet \cite{nam2016learning}, CFNet \cite{valmadre2017end}, SiamFC \cite{bertinetto2016fully} and ECO \cite{danelljan2017eco}.
Results are shown in Table~\ref{tab:trackingnet} for supervised methods and~\ref{tab:ssl_trackingnet} for self-supervised methods.
Again, it can be seen that SiamMask outperforms the competitors according to all metrics considered by the benchmark.
Interestingly, SiamMask even slightly (+2.1\%) improves over ATOM~\cite{danelljan2019atom}, which adapts online the parameters of the network used as a feature extractor.

\begin{table}
% \resizebox{\columnwidth}{!}
% {
\caption{Comparison with unsupervised learning-based methods on the TrackingNet dataset.}
\begin{tabular}{|c|c|c|c|c|}
\hline Metric & LUDT \cite{wang2021unsupervised} & USOT \cite{zheng2021learning} & PUL \cite{wu2021progressive} & SiamMask \\
\hline AUC of success rate & $56.3$ & $61.5$ & $54.6$ & $\mathbf{72.5}$ \\
\hline Tracking precision & $49.5$ & $56.6$ & $48.5$ & $\mathbf{66.4}$ \\
\hline Normalized precision & $53.3$ & $69.1$ & $63.0$ & $\mathbf{77.8}$ \\
\hline
\end{tabular}
% }
\label{tab:ssl_trackingnet}
\end{table}

On the TrackingNet benchmark, we also compared our method with a few unsupervised learning-based visual trackers \cite{wang2021unsupervised,zheng2021learning,wu2021progressive}.
Results are shown in Table~\ref{tab:ssl_trackingnet}.
Unsurprisingly, SiamMask can leverage millions of bounding-box and mask labels during training and achieves significantly better results.

\subsection{Evaluation for video object segmentation (VOS)}
\label{sec:vos_evaluation}

Our model, once trained, can also be used for the task of VOS to achieve competitive performance without requiring any adaptation at test time.
Importantly, differently to typical VOS approaches, ours can operate online, runs in real-time and only requires a simple bounding box initialization.
To initialize SiamMask, an axis-aligned bounding box (the Min-max strategy shown in Fig.~\ref{fig:object_representations}) is extracted from the mask provided in the first frame (multiple objects are tracked and segmented using multiple tracker's instances).
Instead, VOS methods are typically initialized with a binary mask~\cite{perazzi2017video} and many of them require computationally intensive techniques at test time such as fine-tuning~\cite{maninis2018video,perazzi2017learning,bao2018cnn,voigtlaender2017online}, data augmentation~\cite{khoreva2017lucid,li2018video}, inference on MRF/CRF~\cite{wen2015jots,tsai2016video,marki2016bilateral,bao2018cnn} and optical flow~\cite{tsai2016video,bao2018cnn,perazzi2017learning,li2018video,cheng2018fast}.
As a consequence, it is not uncommon for VOS techniques to require several minutes to process a short sequence.
Clearly, these strategies make the online applicability (which is our focus) impossible.
For this reason, in our comparison we mainly concentrate on \emph{fast} VOS approaches.

\begin{figure}[t]
	\centering
	\includegraphics[width=\columnwidth]{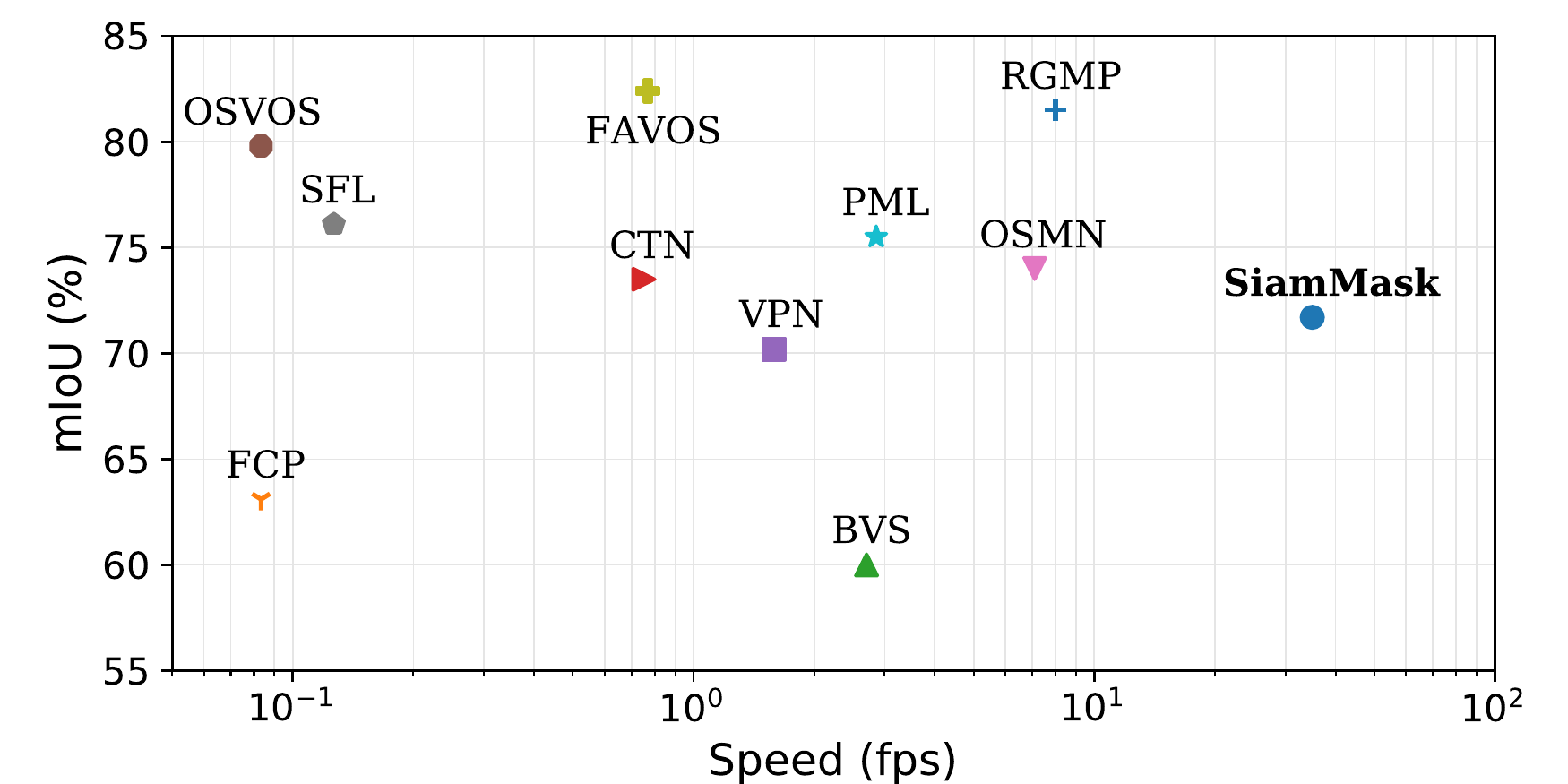}
	\caption{Comparison in terms of mean IOU and speed (fps) between SiamMask and popular fast video object segmentation algorithms on the DAVIS-2016 dataset.}
\label{fig:davis16}
\end{figure}

\noindent\textbf{Results on DAVIS-2016.}
Fig.~\ref{fig:davis16} compares SiamMask with several popular fast VOS methods on DAVIS-2016 in terms of segmentation accuracy (y-axis) and speed (x-axis).
SiamMask shows a comparable accuracy while running significantly faster than other methods (often by an order of magnitude).
Notably, the competitive accuracy is obtained without requiring online updates of the backbone model, as done for instance in OSMN~\cite{yang2018efficient}.

Table~\ref{tab:davis16} offers a more detailed breakdown of the comparison, while also considering slower but better performing methods such as OnAVOS and MSK.
Few notes:
\begin{itemize}
\item OnAVOS~\cite{voigtlaender2017online} and MSK~\cite{perazzi2017learning} yield the best overlap and contour precision across the board.
However, their strategy of performing online model updates make them hundreds of times slower than SiamMask and unable to run in real time.
\item In contrast with VOS approaches which do not perform online model updates (FAVOS\cite{cheng2018fast}, RGMP \cite{oh2018fast}, SFL-ol \cite{cheng2017segflow}, PML \cite{chen2018blazingly}, OSMN \cite{yang2018efficient}, PLM \cite{shin2017pixel} and VPN \cite{jampani2017video}), SiamMask has a simpler initialization (bounding box instead of pixel-wise mask) and an important advantage in terms of speed.
\item Importantly, SiamMask yields the best performance (lowest values) for the decay of both the area overlap rate $\mathcal{J_D}$ and the contour precision $\mathcal{F_D}$).
This suggests that SiamMask is robust over time and thus particularly appropriate to be used in long sequences.
\end{itemize}

\begin{table}
\caption{Comparison between SiamMask and the state-of-the-art segmentation algorithms on the DAVIS 2016 validation set: FT denotes whether fine-tuning is required (\checkmark) or not (\xmark); M denotes whether video segmentation is initialized with a mask (\checkmark) or a bounding box (\xmark); Speed is measured in frames per second.}
    \begin{center}
        \resizebox{.48\textwidth}{!}
        {
        \begin{tabular}{|c|c|c|c|c|c|c|c|c|c|}
        \hline Method & FT & $\mathrm{M}$ & $\mathcal{J_M} \uparrow$ & $\mathcal{J_O} \uparrow$ & $\mathcal{J_D} \downarrow$ & $\mathcal{F_M} \uparrow$ & $\mathcal{F_O} \uparrow$ & $\mathcal{F_D} \downarrow$ & Speed \\
        \hline OnAVOS \cite{voigtlaender2017online} & ${\checkmark}$ & ${\checkmark}$ & $\mathbf{86.1}$ & $96.1$ & $5.2$ & $\mathbf{84.9}$ & $89.7$ & $5.8$ & $0.08$ \\
        \hline MSK \cite{perazzi2017learning} & ${\checkmark}$ & ${\checkmark}$ & $79.7$ & $93.1$ & $8.9$ & $75.4$ & $87.1$ & $9.0$ & $0.1$ \\
        \hline MSK $_{b}$\cite{perazzi2017learning} & ${\checkmark}$ & \xmark & $69.6$ & $-$ & $-$ & $-$ & $-$ & $-$ & $0.1$ \\
        \hline SFL \cite{cheng2017segflow} & ${\checkmark}$ & ${\checkmark}$ & $76.1$ & $90.6$ & $12.1$ & $76.0$ & $85.5$ & $10.4$ & $0.1$ \\
        \hline FAVOS \cite{cheng2018fast} & \xmark & ${\checkmark}$ & $82.4$ & $\mathbf{96.5}$ & $4.5$ & $79.5$ & $89.4$ & $5.5$ & $0.8$ \\
        \hline RGMP \cite{oh2018fast} & \xmark & ${\checkmark}$ & $81.5$ & $91.7$ & $10.9$ & $82.0$ & $\mathbf{90.8}$ & $10.1$ & 8 \\
        \hline SFL-ol \cite{cheng2017segflow} & \xmark & ${\checkmark}$ & $67.4$ & $81.4$ & $6.2$ & $66.7$ & $77.1$ & $5.1$ & $3.3$ \\
        \hline PML \cite{chen2018blazingly} & \xmark & ${\checkmark}$ & $75.5$ & $89.6$ & $8.5$ & $79.3$ & $93.4$ & $7.8$ & $3.6$ \\
        \hline OSMN \cite{yang2018efficient} & \xmark & ${\checkmark}$ & $74.0$ & $87.6$ & $9.0$ & $72.9$ & $84.0$ & $10.6$ & $8.0$ \\
        \hline PLM \cite{shin2017pixel} & \xmark & ${\checkmark}$ & $70.2$ & $86.3$ & $11.2$ & $62.5$ & $73.2$ & $14.7$ & $6.7$ \\
        \hline VPN \cite{jampani2017video} & \xmark & ${\checkmark}$ & $70.2$ & $82.3$ & $12.4$ & $65.5$ & $69.0$ & $14.4$ & $1.6$ \\
        \hline SiamMask & \xmark & \xmark & $71.7$ & $86.8$ & $\mathbf{3.0}$ & $67.8$ & $79.8$ & $\mathbf{2.1}$ & $\mathbf{55}$ \\
        \hline
        \end{tabular}
        }
    \end{center}
\label{tab:davis16}
\end{table}

\newpage
\noindent\textbf{Results on DAVIS-2017 and YouTube-VOS.}

\begin{table}
\caption{Comparison on the DAVIS 2017 validation set.}
    \begin{center}
        \resizebox{.48\textwidth}{!}{
            \begin{tabular}{|c|c|c|c|c|c|c|c|c|c|}
            \hline Method & FT & M & $\mathcal{J_M} \uparrow$ & $\mathcal{J_O} \uparrow$ & $\mathcal{J_D} \downarrow$ & $\mathcal{F_M} \uparrow$ & $\mathcal{F_O} \uparrow$ & $\mathcal{F_D} \downarrow$ & Speed \\
            \hline OnAVOS \cite{voigtlaender2017online} & ${\checkmark}$ & ${\checkmark}$ & $\mathbf{6 1 . 6}$ & $\mathbf{6 7 . 4}$ & $27.9$ & $\mathbf{6 9 . 1}$ & $\mathbf{7 5 . 4}$ & $26.6$ & $0.1$ \\
            \hline OSVOS \cite{caelles2017one} & ${\checkmark}$ & ${\checkmark}$ & $56.6$ & $63.8$ & $26.1$ & $63.9$ & $73.8$ & $27.0$ & $0.1$ \\
            \hline FAVOS \cite{cheng2018fast} & \xmark & ${\checkmark}$ & $54.6$ & $61.1$ & $\mathbf{1 4 . 1}$ & $61.8$ & $72.3$ & $\mathbf{1 8 . 0}$ & $0.8$ \\
            \hline OSMN \cite{yang2018efficient} & \xmark & ${\checkmark}$ & $52.5$ & $60.9$ & $21.5$ & $57.1$ & $66.1$ & $24.3$ & $8.0$ \\
            \hline SiamMask & \xmark & \xmark & $54.3$ & $62.8$ & $19.3$ & $58.5$ & $67.5$ & $20.9$ & 55 \\
            \hline
            \end{tabular}
        }
    \end{center}
\label{tab:davis2017}
\end{table}

Table~\ref{tab:davis2017} and~\ref{tab:youtube_vos} compare the performance of SiamMask on two additional VOS benchmarks: DAVIS-2017 and YouTube-VOS.
Looking at the results from the tables, few comments can be made:
\begin{itemize}
    \item SiamMask again does not have the best performance overall, but it is very competitive at a speed that is often hundreds of time faster than higher performing methods like OnAVOS and OSVOS.
    \item On DAVIS-2017, again SiamMask exhibits a strong temporal robustness (expressed by a low decay). It is surpassed only by the slower FAVOS, which maintains several trackers for different object parts, thus being able to deal with complex deformations that naturally occur over time.
    \item On YouTube-VOS, SiamMask surprisingly achieves the best accuracy for the set of \emph{seen} classes.
    \item The second fastest method after SiamMask is OSMN, which uses meta learning to perform rapid online parameters updates. However, it performs worse across the board in all metrics.
\end{itemize}

\begin{table}
    \caption{Comparison on the YouTube-VOS validation set.}
    \begin{center}
        \resizebox{.48\textwidth}{!}
        {
            \begin{tabular}{|c|c|c|c|c|c|c|c|c|}
            \hline Method & FT & M & $\mathcal{J_S}$ $\uparrow$ & $\mathcal{J_U}$ $\uparrow$ & $\mathcal{F_S}$ $\uparrow$ & $\mathcal{F_U}$ $\uparrow$ & $\mathcal{O}$ $\uparrow$ & Speed $\uparrow$ \\
            \hline OnAVOS \cite{voigtlaender2017online} & ${\checkmark}$ & ${\checkmark}$ & $60.1$ & $46.6$ & $\mathbf{62.7}$ & $51.4$ & $55.2$ & $0.1$ \\
            \hline OSVOS \cite{caelles2017one} & ${\checkmark}$ & ${\checkmark}$ & $59.8$ & $\mathbf{54.2}$ & $60.5$ & $\mathbf{60.7}$ & $\mathbf{58.8}$ & $0.1$ \\
            \hline OSMN \cite{yang2018efficient} & \xmark & ${\checkmark}$ & $60.0$ & $40.6$ & $60.1$ & $44.0$ & $51.2$ & $8.0$ \\
            \hline SiamMask & \xmark & \xmark & $\mathbf{60.2}$ & $45.1$ & $58.2$ & $47.7$ & $52.8$ & $55$ \\
            \hline
            \end{tabular}
        }
    \end{center}
\label{tab:youtube_vos}
\end{table}

\smallskip\noindent\textbf{General remarks.}
The results on DAVIS-2016, DAVIS-2017 and YouTube-VOS (Table~\ref{tab:davis16}, \ref{tab:davis2017} and~\ref{tab:youtube_vos}) demonstrate that SiamMask can achieve competitive online segmentation performance at a fast speed, with only a simple bounding-box initialization and without any adaptation at test time.
Moreover, SiamMask \emph{(1)} is almost two orders of magnitude faster than accurate segmentation algorithms such as OnAVOS \cite{voigtlaender2017online} and SFL \cite{cheng2017segflow}; \emph{(2)} is about four times faster than fast online methods such as OSMN \cite{yang2018efficient} and RGMP \cite{oh2018fast}; \emph{(3)} has a remarkably low decay in performance over time, which makes it effective in long sequences.
These points suggest that SiamMask can be treated as a strong baseline for online video object segmentation, as well as tracking.

\subsection{Evaluation for multiple object tracking and segmentation}
\label{sec:mot_evaluation}

First, we validate the efficacy of the two-stage strategy when performing multiple object tracking and segmentation on the validation set of YouTubeVIS~\cite{yang2019video}.
The comparison of the one-stage vs. the two-stage approach is shown in Table~\ref{tab:vis_1vs2stages}.
In both cases, one tracker per target object is instantiated and the off-the-shelf segmentation approach HTC~\cite{chen2019hybrid} is used to initialize the tracks.
As it can be seen, the two-stage variant modestly improves the results, with an absolute $1.5\%$ improvement in mAP and $1.6\%$ in average recall.
We believe this improvement is to attribute to the fact that the two-stage version regression branch helps limiting the area to segment, thus reducing the degree of the difficulty for the mask branch.

\begin{table}
    \caption{Comparison between the proposed two-stage SiamMask and the single stage SiamMask on the validation set of YouTubeVIS~\cite{yang2019video}. mAP refers to mean average precision, while $\mathrm{AR}@10$ is the average recall with 10 proposals, averaged over IoU thresholds and classes.
    $d_{mAP}$ and $d_{AR10}$ indicate the difference for both metrics. HTC is the off-the-shelf segmentation method~\cite{chen2019hybrid} used to initialize the tracks.}
    \begin{center}
        \resizebox{.48\textwidth}{!}{
            \begin{tabular}{|c|c|c|c|c|}
            \hline Model & $\mathrm{mAP}$ & $d_{\mathrm{mAP}}$ & $\mathrm{AR} @ 10$ & $d_{\mathrm{AR} 10}$ \\
            \hline HTC+SiamMask & $0.366$ & $--$ & $0.423$ & $--$ \\
            \hline HTC+ two-stage SiamMask & $0.381$ & $+0.015$ & $0.439$ & $+0.016$ \\
            \hline
            \end{tabular}
        }
    \end{center}
\label{tab:vis_1vs2stages}
\end{table}

\begin{table}
    \begin{center}
    \caption{Leaderboard of the 2019 edition of YouTube-VIS. The metrics used by the challenge are mean average precision (mAP), average precision at fixed IOU threshold of 50\% and 75\% (AP50 and AP75), and average recall with 1 or 10 proposals (AR1 and AR10).
    For more information, and for a description of the all the approaches used in the competition, see~\url{https://youtube-vos.org/challenge/2019/leaderboard}.}
        \resizebox{.48\textwidth}{!}{
           \begin{tabular}{|c|c|c|c|c|c|c|}
            \hline Team name & mAP & AP50 & AP75 & AR1 & AR10 & Ranking\\
            \hline Jono \cite{luiten2019video} & $\mathbf{0.467}$ & $\mathbf{0.697}$ & $\mathbf{0.509}$ & $\mathbf{0.462}$ & $\mathbf{0.537}$ & 1 \\
            \hline \emph\textbf{{Ours}} & $0.457$ & $0.674$ & $0.490$ & $0.435$ & $0.507$ & 2\\
            \hline bellejuillet \cite{feng2019dual} & $0.450$ & $0.636$ & $0.502$ & $0.447$ & $0.503$ & 3\\
            \hline Linhj & $0.449$ & $0.665$ & $0.486$ & $0.453$ & $0.538$ & 4\\
            \hline mingmingdiii \cite{dong2019temporal} & $0.444$ & $0.684$ & $0.487$ & $0.436$ & $0.508$ & 5\\
            \hline xiAaonice \cite{liu2019spatio} & $0.400$ & $0.578$ & $0.449$ & $0.396$ & $0.452$ & 6\\
            \hline guwop  & $0.400$ & $0.608$ & $0.439$ & $0.412$ & $0.491$ & 7\\
            \hline exing & $0.397$ & $0.621$ & $0.426$ & $0.414$ & $0.461$ & 8\\
            \hline \emph{Baseline}~\cite{jampani2017video} & $0.313$ & $0.503$ & $0.338$ & $0.335$ & $0.369$ & --\\
            \hline
            \end{tabular}
        }
    \label{tab:youtubevis}
    \end{center}
\end{table}

Table~\ref{tab:youtubevis} shows the comparison between the two-stage version of SiamMask and the algorithms participating to the 2019 YouTube-VIS challenge (on the test-set)~\cite{yang2019video}.
Compared with the official baseline proposed by the YouTube-VIS organizers~\cite{yang2019video}, the two-stage version of SiamMask increases the mAP of a relative $46\%$.
Despite the very simple approach, SiamMask ranks second on the leaderboard, after the approach described in~\cite{luiten2019video}, which consider the VIS task as constituted by four different problems: detection, classification, segmentation, and tracking, and solves them separately.

One important situation that our simple adaptation to SiamMask does not handle well is \emph{confusion}: when multiple objects are very close to each other, there is a high uncertainty in mapping pixels to identities, with the results that one object can ``hijack'' the mask of another.
A solution to this issue could be to model the relationship between pixels within the same mask (\emph{e.g.} with conditional random fields, or graph neural networks), rather than treating each pixel individually.
However, this will inevitably reduce the tracking speed.

\subsection{Ablation studies}
\label{sec:ablations}
We perform a series of ablation studies to analyse the impact of different architectures and multi-task training setups.

Table~\ref{tab:ablations} compares different variants of the fully-convolutional Siamese framework, indicating whether the classic AlexNet or ResNet-50 are used as  the backbone $f_{\theta}$, whether or not the mask refinement strategy (from~\cite{o2015learning}) is used, and which multi-task configuration is adopted.
Few observations can be made:
\begin{itemize}
    \item Unsurprisingly, using a ResNet-50 backbone delivers better performance, and at a reasonable cost in terms of speed.
    \item Using the same ResNet-50 backbone, the two- and three-branch variants of SiamMask improve over their respective baselines, SiamFC and SiamRPN.
    \item Mask refinement is very useful for increasing the contour accuracy in the segmentation task. However, it does not seem to significantly affect the EAO tracking metric. This is not particularly surprising, as it only considers rotated bounding boxes, which are only a crude approximation of the actual object boundaries.
\end{itemize}

\begin{table}
    \caption{Ablation studies of SiamMask on the VOT-2018 and DAVIS-2016 datasets}
    \begin{center}
        \resizebox{.48\textwidth}{!}{
            \begin{tabular}{|c|c|c|c|c|c|c|}
            \hline Method & AlexNet & ResNet-50 & EAO $\uparrow$ & $\mathcal{J_M} \uparrow$ & $\mathcal{F_M} \uparrow$ & Speed (fps) \\
            \hline SiamFC & ${\checkmark}$ & & $0.188$ & $-$ & $-$ & 86 \\
            \hline SiamFC & & ${\checkmark}$ & $0.251$ & $-$ & $-$ & 40 \\
            \hline SiamRPN & ${\checkmark}$ & & $0.243$ & $-$ & $-$ & $\mathbf{2 0 0}$ \\
            \hline SiamRPN & & ${\checkmark}$ & $0.359$ & $-$ & $-$ & 76 \\
            \hline SiamMask-2 branches without mask refinement & & ${\checkmark}$ & $0.326$ & $62.3$ & $55.6$ & 43 \\
            \hline SiamMask-3braches without mask refinement & & ${\checkmark}$ &$0.375$ & $68.6$ & $57.8$ & 58 \\
            \hline SiamMask-2branches-score & & ${\checkmark}$ & $0.265$ & $-$ & $-$ & 40 \\
            \hline SiamMask-3branches-box & & ${\checkmark}$ & $0.363$ & $-$ & $-$ & 76 \\
            \hline SiamMask-2braches & & ${\checkmark}$ & $0.334$ & $67.4$ & $63.5$ & 60 \\
            \hline SiamMask-3branches & & ${\checkmark}$ & $\mathbf{0 . 3 8 0}$ & $\mathbf{71.7}$ & $\mathbf{6 7 . 8}$ & 55 \\
            \hline
            \end{tabular}
        }
    \end{center}
\label{tab:ablations}
\end{table}

We conducted two further experiments to disentangle the effect of multi-task training, also reported in Table~\ref{tab:ablations}.
To achieve this, we modified the two variants of SiamMask during inference so that, respectively, they report an axis-aligned bounding box from the score branch (SiamMask-2branches-score) or the box branch (SiamMask-3branches-box).
Therefore, despite having been trained, the mask branch is \emph{not} used during inference.
We can observe how both variants obtain a modest but meaningful improvement with respect to their counterparts (SiamFC and SiamRPN): from 0.251 to 0.265 EAO for the \emph{two-branch} and from 0.359 to 0.363 for the \emph{three-branch}.
This might suggests that learning an additional task could act as a regularizer even when the segmentation output is not used, although our experimental setup is too limited to draw such a conclusion with confidence.

\subsection{Qualitative examples}
\label{sec:qualitative_examples}

\begin{figure}[t]
	\centering
	\includegraphics[width=0.9\columnwidth]{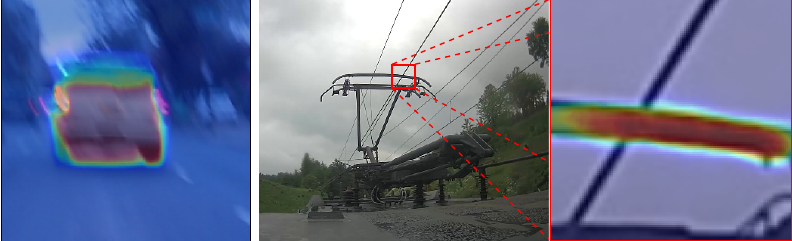}
	\caption{Failure cases: extreme motion blur and ``non-object'' initialization.}
\label{fig:failures}
\end{figure}

\begin{figure}[t]
	\centering
	\includegraphics[width=\columnwidth]{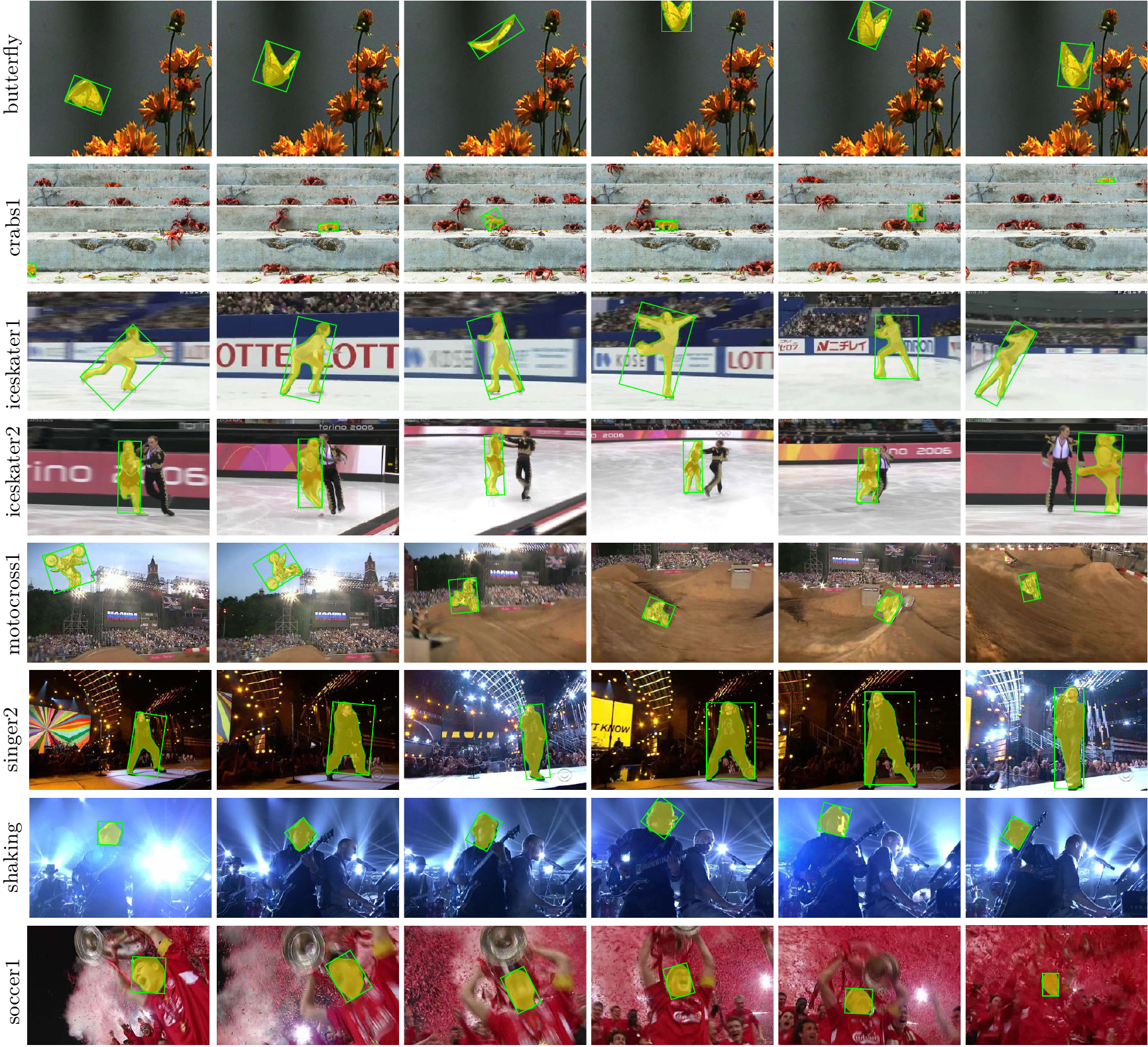}
	\caption{Qualitative results of SiamMask for the sequences \texttt{butterfly}, \texttt{crabs1}, \texttt{iceskater1}, \texttt{iceskater2}, \texttt{motocross1}, \texttt{singer2}, \texttt{shaking}, and \texttt{soccer1} from the visual object tracking benchmark VOT-2018.}
\label{fig:vot_qualitative}
\end{figure}

\smallskip\noindent\textbf{Failure cases.}
\label{sec:qualitative_results}
Qualitatively, we observed a few scenarios in which SiamMask performs rather poorly.
One is extreme motion blur (\emph{e.g.} left-hand side of Fig.~\ref{fig:failures}), which is caused by sudden camera motion or by the fast movements of the target object.
Since labelling is itself expected to present a significant amount of noise, this is a scenario where a supervised, offline-trained-only strategy like SiamMask can be particularly susceptible.
Instead, when motion blur is not extreme, SiamMask typically perform rather well (see \emph{e.g.} some of the examples in Fig.~\ref{fig:vot_qualitative}.

As already mentioned earlier when considering the multiple-object case, since SiamMask models pixels individually, \textit{confusion} (\emph{i.e.} when different object's trajectories overlap) is another very challenging scenario, even if we are only interested in one target.

Finally, a rather pathological but still important failure case is the one encountered when the selected area to track does not correspond to an object, but rather to a texture or a part of an object (\emph{e.g.} see the right part of Fig.~\ref{fig:failures}).
Given that SiamMask is trained on a large dataset with object-level labels, it is naturally biased towards objects even when the initialization provided by the user says otherwise.

\begin{figure}[t]
	\centering
	\includegraphics[width=0.9\columnwidth]{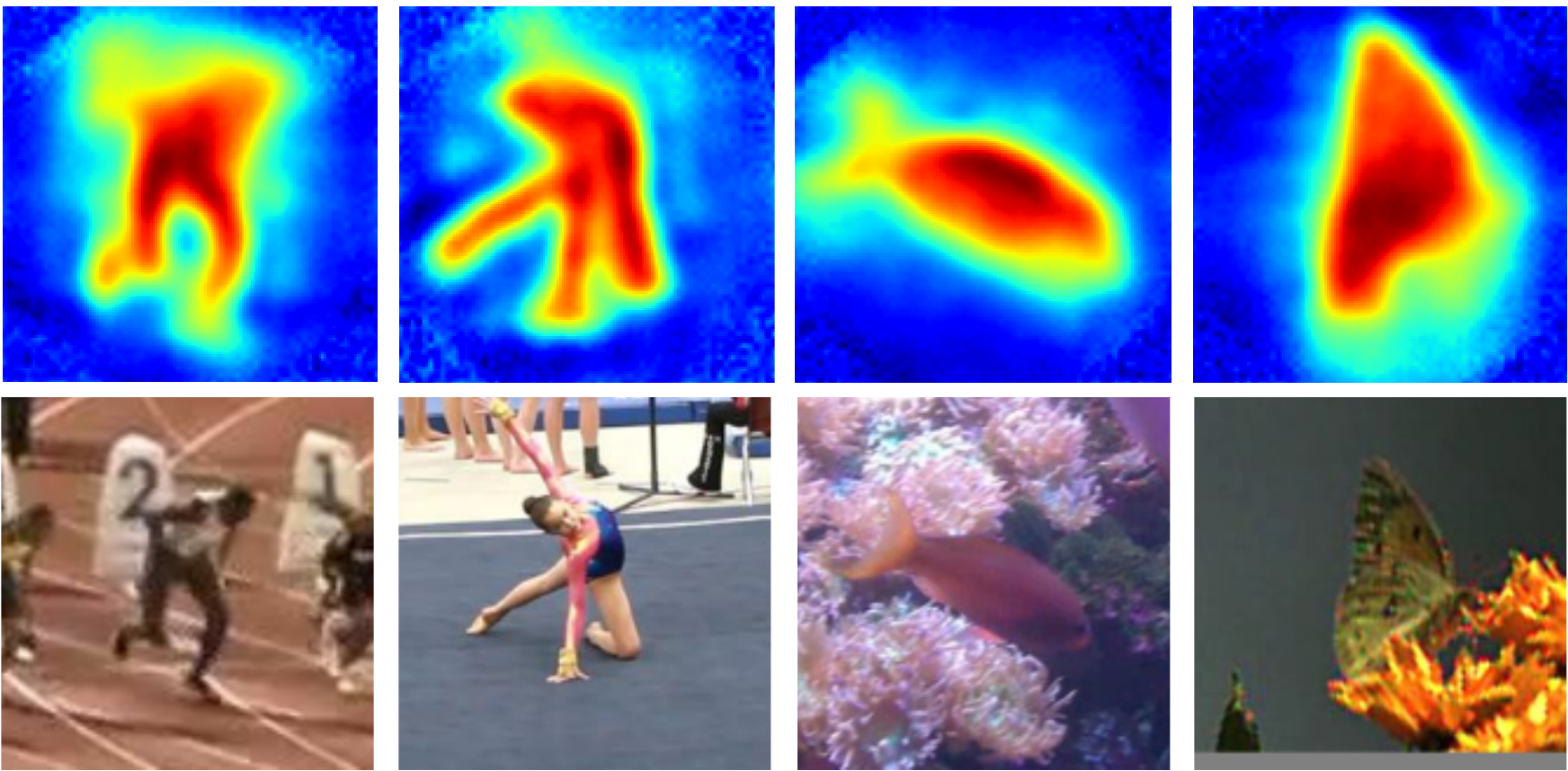}
	\caption{Example score maps from the mask branch for a variety of objects.}
\label{fig:shapes}
\end{figure}

\smallskip\noindent\textbf{A variety of objects and shapes.}
Fig.~\ref{fig:shapes} shows a few qualitative examples of mask predictions for objects of different types and shapes.
In general, we observe that SiamMask adapts well to all sort of objects and deformations, and provides fairly accurate masks even in presence of noisy backgrounds and non-rigid deformations.

\begin{figure}[t]
	\centering
	\includegraphics[width=0.9\columnwidth]{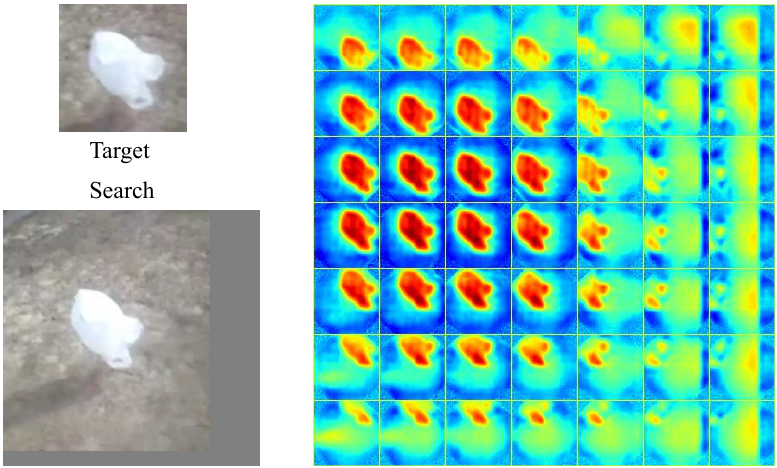}
	\caption{Score maps from the mask branch at different locations.}
    \label{fig:multiple_masks_visualization}
\end{figure}

\smallskip\noindent\textbf{Multiple masks per output.}
SiamMask generates one mask for each individual RoW (response of a candidate window).
During tracking, the RoW attaining the maximum score from the classification branch is considered as the region in which the object is.
The segmentation branch predicts the final output mask corresponding to this region. To more clearly observe what the mask branch predicts, we visualize the masks predicted from different RoWs of the same search area in Fig.~\ref{fig:multiple_masks_visualization}.

\smallskip\noindent\textbf{Further qualitative results.}
 In order to qualitatively analyze tracking and segmentation accuracy of SiamMask, we present the visual results of SiamMask for some challenging video sequences from the VOT-2018, DAVIS-2016 and 2017, and Youtube-VIS datasets.

Sequences for the single-object tracking benchmark VOT-2018 are shown in Fig.\ref{fig:vot_qualitative}.
SiamMask maintains high accuracy with sequences presenting important non-rigid deformations, like \texttt{butterfly} and \texttt{iceskater1}.
While \texttt{butterfly} is a fairly ``simple'' sequence because of the stark contrast between object and background, \texttt{iceskater1} is challenging, because of the complexity of the background.
With fast-moving objects, SiamMask can produce accurate segmentation masks even in presence of distractors (\emph{e.g} \texttt{crabs1} and \texttt{iceskater2}). However, as seen previously, conditions become much more challenging for SiamMask when object trajectories overlap.
Video object segmentation algorithms are often sensitive to motion blur and variations in illumination.
On the contrary, SiamMask yields accurate masks for the sequences \texttt{singer2}, \texttt{shaking}, and \texttt{soccer1}, which present large variations in illumination and severe motion blur.

\begin{figure}[t]
	\centering
	\includegraphics[width=\columnwidth]{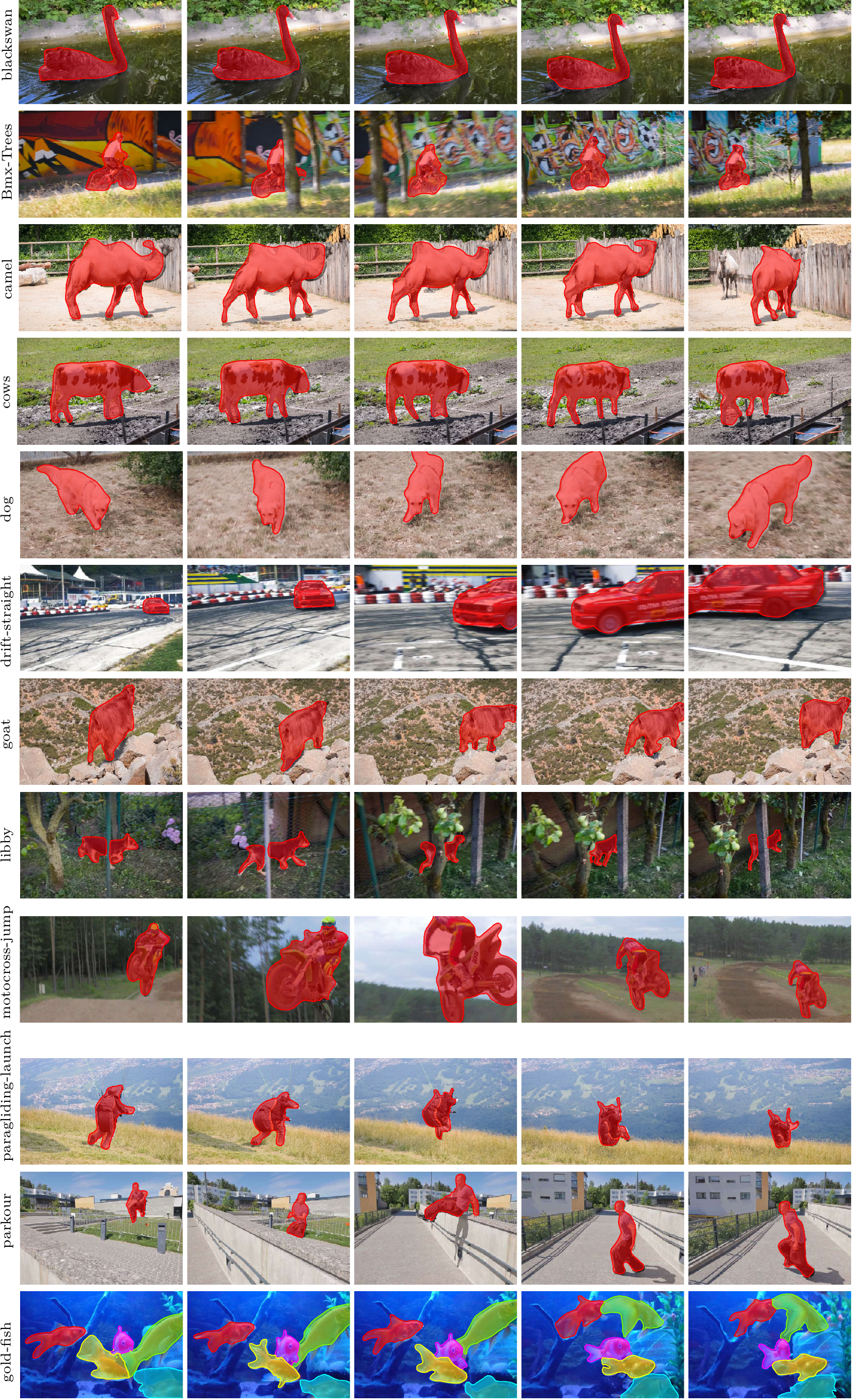}
	\caption{Qualitative results of SiamMask for some sequences from the object segmentation benchmarks DAVIS-2016 and DAVIS-2017.}
\label{fig:vos_qualitative}
\end{figure}

Fig.~\ref{fig:vos_qualitative} shows the qualitative results of SiamMask on a few representative sequences from DAVIS-2016 and DAVIS-2017.
For the video object segmentation task, SiamMask effectively adapts to the challenge presented by change of scales, view angle, and shape (\emph{e.g.} \texttt{drift-straight} and \texttt{motocross-jump} sequences).
Moreover, it is also effective in accurately handling minor occlusion of the target object (\emph{e.g.} \texttt{bmx-trees} and \texttt{libby} sequences).

\begin{figure}[!t]
	\centering
	\includegraphics[width=\columnwidth]{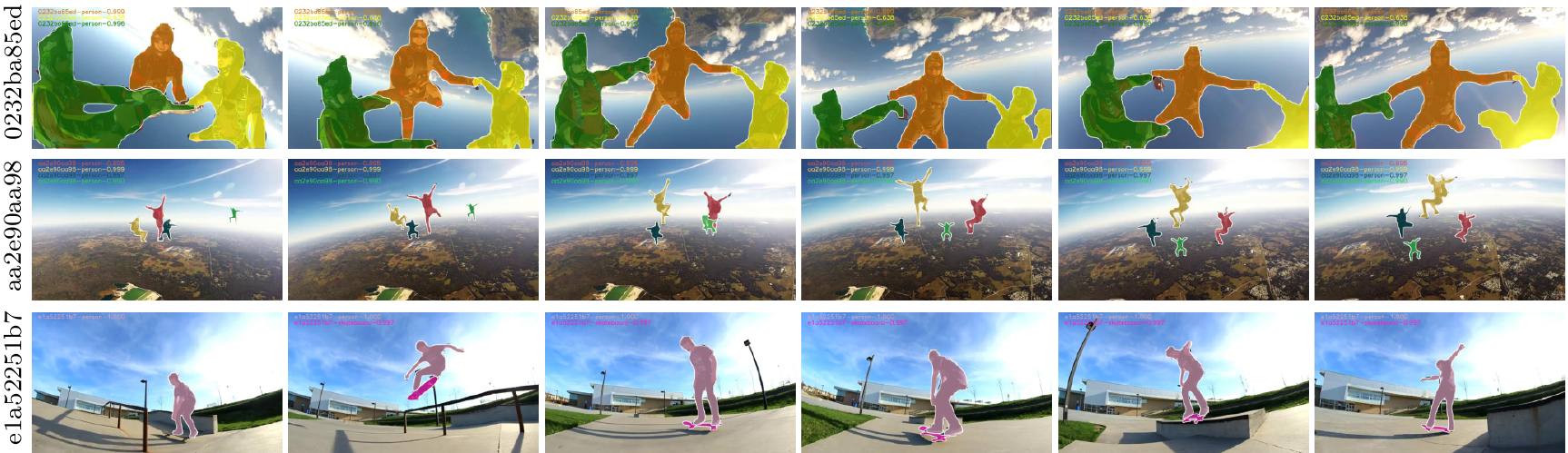}
	\caption{Example results of the two-stage version of SiamMask on some sequences from Youtube-VIS.}
	\vspace{-0.5em}
\label{fig:vis_qualitative}
\end{figure}

Finally, Fig.~\ref{fig:vis_qualitative} visualizes a set of positive results of our two-stage SiamMask variant described in~\ref{sec:MOT} on some challenging videos from YouTube-VIS, where objects undergo deformation, occlusion, or rapid motion.

\section{Conclusion}
\label{sec:conclusion}
In this paper we introduced SiamMask, a simple approach that enables fully-convolutional Siamese trackers to produce class-agnostic binary segmentation masks of the target object.
We show how it can be applied with success to both tasks of visual object tracking \emph{and} semi-supervised video object segmentation, showing better accuracy than most real-time trackers and, at the same time, the fastest speed among VOS methods.
The two variants of SiamMask we proposed are initialised with a simple bounding box, operate online, run in real-time and do not require any adaptation to the test sequence.
In addition, SiamMask can be easily extended to also perform multiple object tracking and segmentation by cascading two models.
We hope that our work will inspire further studies on multi-task approach that consider different but closely related computer vision problems together.

% \newpage

\bibliographystyle{IEEEtran}

\bibliography{bibliography}

%\clearpage

\begin{IEEEbiography}[{\includegraphics[width=1in,height=1.25in,clip,keepaspectratio]{mshell}}]{Michael Shell}
\end{IEEEbiography}

\begin{wrapfigure}{l}{1.2cm}
\includegraphics[width=1.2cm]{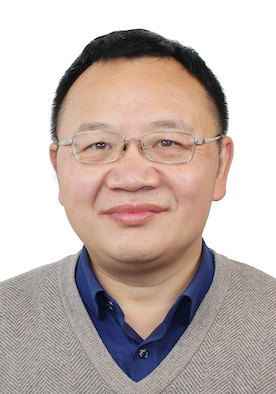}
\end{wrapfigure}
\noindent \textbf{Weiming Hu} received the Ph.D. degree from the department of computer science and engineering, Zhejiang University in 1998. From April 1998 to March 2000, he was a postdoctoral research fellow with the Institute of Computer Science and Technology, Peking University. Now he is a professor in the Institute of Automation, Chinese Academy of Sciences. His research interests are in visual motion analysis, recognition of web objectionable information, and network intrusion detection.

\vspace{0.5cm}
\begin{wrapfigure}{l}{1.2cm}
\includegraphics[width=1.2cm]{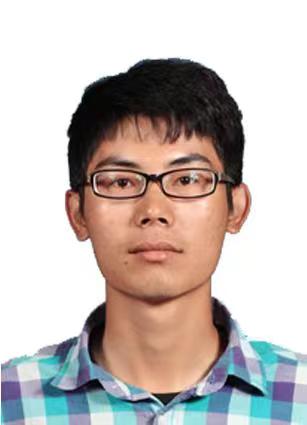}
\end{wrapfigure}
\noindent \textbf{Qiang Wang} received the B.S. degree in automation from the University of Science and Technology Beijing, Beijing, China, in 2015. He is currently pursuing the Ph.D. degree with the Institute of Automation, University of Chinese Academy of Sciences (UCAS). His research interest includes the theory and applications of single object tracking.

\vspace{0.5cm}
\begin{wrapfigure}{l}{1.2cm}
\includegraphics[width=1.2cm]{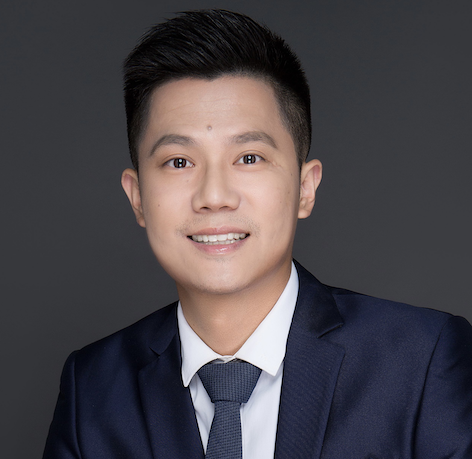}
\end{wrapfigure}
\noindent \textbf{Li Zhang}
is a tenure-track Professor at Fudan University.
Previously, he was a Postdoctoral Research Fellow at the University of Oxford.
Prior to joining Oxford, he read his PhD in computer science at Queen Mary University of London.
His research interests include computer vision and deep learning.

\vspace{0.5cm}
\begin{wrapfigure}{l}{1.2cm}
\includegraphics[width=1.2cm]{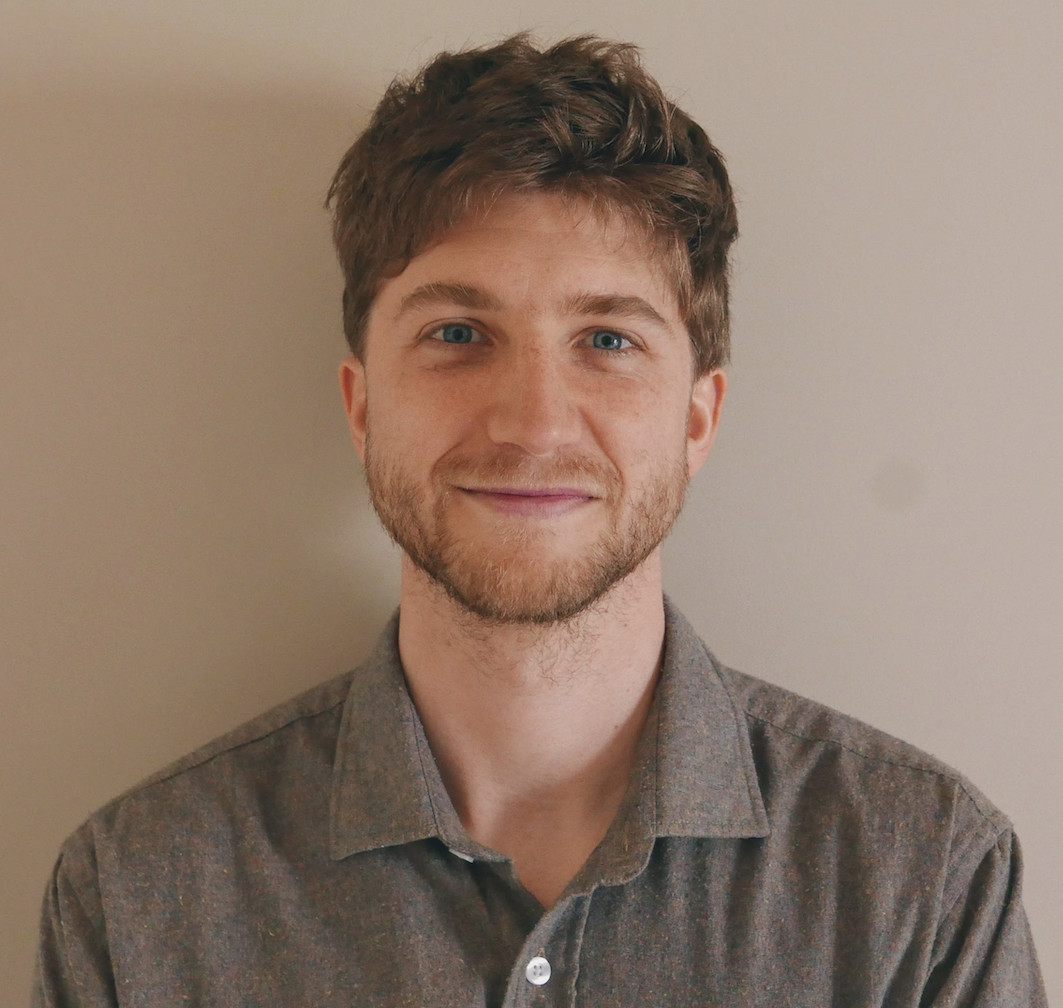}
\end{wrapfigure}
\noindent \textbf{Luca Bertinetto} completed his PhD at Oxford University in 2019 with the thesis \emph{``Learning (to learn) with limited data''}, during which he published several influential works on the topics of object tracking and meta-learning.
Currently, he is a senior research scientist at Five, where he is particularly interested in improving the robustness of machine learning systems in realistic scenarios for which supervision is limited (or absent), and data does not necessarily resemble the training set.

\vspace{0.5cm}
\begin{wrapfigure}{l}{1.2cm}
\includegraphics[width=1.2cm]{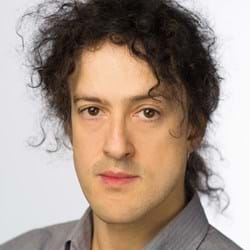}
\end{wrapfigure}
\noindent \textbf{Philip Torr} received the PhD degree from Oxford University. After working for another three years at Oxford, he worked for six years for Microsoft Research, first in Redmond, then in Cambridge, founding the vision side of the Machine Learning and Perception Group. He is now a professor at Oxford University. He has won awards from top vision conferences, including ICCV, CVPR, ECCV, NeurIPS and BMVC. He is a senior member of the IEEE and a Royal Society Wolfson Research Merit Award holder.

\end{document}